\documentclass[letterpaper, 10 pt, conference]{ieeeconf}  

\IEEEoverridecommandlockouts                              

\usepackage{graphicx} 
\usepackage{hyperref}
\usepackage{todonotes}
\usepackage{amssymb}
\usepackage{amsmath}
\usepackage{svg}
\usepackage[caption=false,font=footnotesize]{subfig}
\usepackage{censor}
\title{\LARGE \bf
GHOST: Ground-projected Hypotheses from Observed Structure-from-Motion Trajectories
}

\author{Tomasz Frelek*, Rohan Patil*, Akshar Tumu* and Henrik I. Christensen \\ Department of Computer Science and Engineering, UC San Diego, United States
\thanks{*Equal Contribution}
}

\begin{document}

\maketitle
\thispagestyle{empty}
\pagestyle{empty}

\begin{abstract}

We present a scalable self-supervised approach for segmenting feasible vehicle trajectories from monocular images for autonomous driving in complex urban environments. Leveraging large-scale dashcam videos, we treat recorded ego-vehicle motion as implicit supervision and recover camera trajectories via monocular structure-from-motion, projecting them onto the ground plane to generate spatial masks of traversed regions without manual annotation. These automatically generated labels are used to train a deep segmentation network that predicts motion-conditioned path proposals from a single RGB image at run time, without explicit modeling of road or lane markings. Trained on diverse, unconstrained internet data, the model implicitly captures scene layout, lane topology, and intersection structure, and generalizes across varying camera configurations. We evaluate our approach on NuScenes, demonstrating reliable trajectory prediction, and further show transfer to an electric scooter platform through light fine-tuning. Our results indicate that large-scale ego-motion distillation yields structured and generalizable path proposals beyond the demonstrated trajectory, enabling trajectory hypothesis estimation via image segmentation. 
\end{abstract}

\section{Introduction}
\label{sec:introduction}
    Understanding where a vehicle can and will move is a fundamental problem in autonomous driving. Classical approaches rely on explicit semantic supervision, costly human annotations, or calibrated sensor suites to learn drivable space and motion priors. However, large-scale dashcam video repositories already contain vast amounts of implicit supervision: every video encodes a real vehicle's interaction with the environment. The ego-vehicle's trajectory is, in effect, a demonstration of feasible motion conditioned on scene geometry, layout, traffic rules, and social constraints.

    In this work, we propose leveraging this implicit supervision by distilling ego-motion directly from monocular dashcam videos. Given raw driving footage, we recover camera trajectories using monocular structure-from-motion (SfM) techniques. These trajectories are projected onto the ground plane to generate spatial masks representing the region traversed by the ego vehicle.
    
    We train a segmentation model to predict these ground-projected trajectory masks directly from a single RGB image. Unlike traditional drivable-area segmentation, our supervision is not manually annotated and does not explicitly encode semantic labels. Instead, it reflects real human driving behavior embedded in large scale unconstrained internet videos. The learned model implicitly captures scene semantics, lane topology, curvature, and intersection structure. Furthermore, as the data comes from a variety of different cameras that have different intrinsic and extrinsic parameters, the model is able to generalize to these various settings.  We also observe that the predicted masks extend beyond the exact demonstrated trajectory, often covering alternative feasible routes, suggesting that the model learns a structured prior over possible motion conditioned on scene appearance.
    
    We evaluate our method on an annotated dataset (NuScenes~\cite{caesar2020nuscenes}) to demonstrate the efficacy of the approach. In addition, we fine-tune the trained model on data collected from an electric scooter to showcase that we can use the same approach along with the collected large scale data to transfer to other vehicles as well. We believe this will be of particular interest to people who work on the autonomy of vehicles other than cars, where structured datasets are sparse, and the drivable area doesn't necessarily equate to traditional road structures.
    
    In summary, our contributions are:

    \begin{enumerate}
        \item A scalable self-supervised pipeline extracts ego-motion supervision from unconstrained monocular dashcam videos using structure-from-motion.
        \item Formulate the problem of estimating future trajectory hypotheses as a segmentation problem.
        \item Empirical evidence that at scale supervision induces generalizable and structured path predictions beyond the demonstrated trajectory using a standard segmentation model.
        \item Transfer the trained model to an electric scooter, and establish that the model effectively generalizes to diverse trajectory estimation via image segmentation.
    \end{enumerate}

    \begin{figure*}[t]
        \centering
        
        \includegraphics[width=0.7\linewidth]{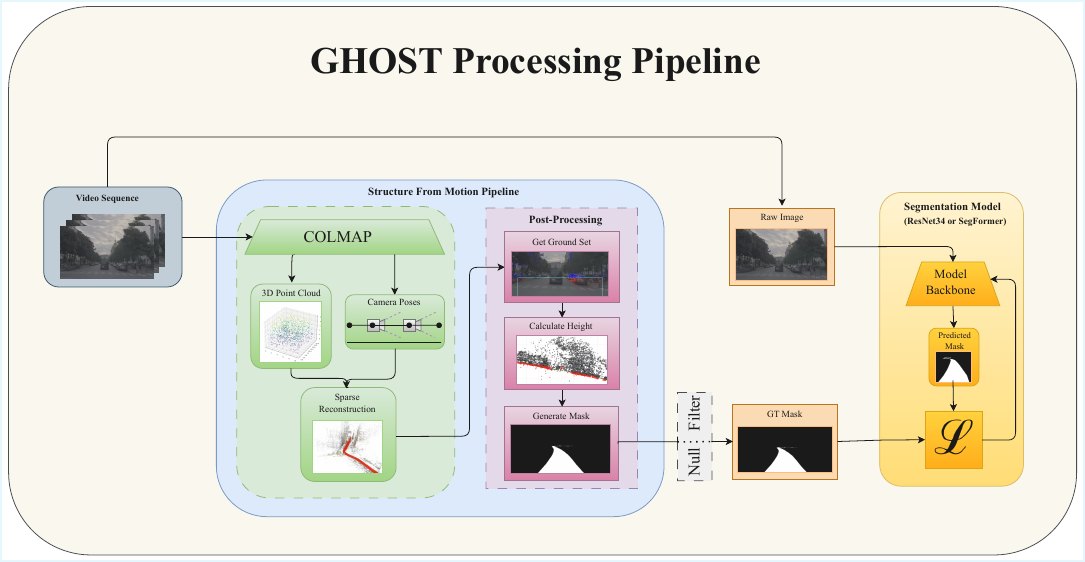}
        \caption{\textbf{Overview of the GHOST self-supervised trajectory estimation pipeline.} The framework consists of three main stages: (1) SfM Pipeline: Raw uncalibrated video sequences are processed using COLMAP to extract sparse 3D point clouds and unscaled 6-DoF camera poses. (2) Post-Processing \& Mask Generation: Ground points are recovered from the sequence's sparse reconstruction; camera height is calculated; recovered height is used to project the trajectory onto the ground plane, generating spatially accurate Ground Truth (GT) masks. (3) Model Training: A segmentation model takes a batch of raw RGB images as input and predicts probabilistic spatial masks. The model is trained by calculating a weighted asymmetric loss $\mathcal{L}$~\ref{eq:loss} between the predicted hypotheses and the auto-generated GT masks.}
        \label{fig:Pipeline}
    \end{figure*}
\section{Related Works}
\label{sec:related}
    Our approach to learning feasible vehicle motion sits at the intersection of self-supervised trajectory estimation and image-based scene understanding. Because monocular camera pose estimation inherently suffers from scale ambiguity~\cite{schoenberger2016sfm}, we formulate trajectory prediction as an image-space segmentation task. Consequently, our review of the literature focuses primarily on image segmentation-based navigation methods, with further elaboration on scale constraints in Section~\ref{subsec:scale_ambiguity}.
    
    \subsection{Monocular Structure from Motion}

        Monocular SfM estimates camera poses and reconstructs 3D scene structure from image sequences. A widely adopted classical pipeline is COLMAP~\cite{schoenberger2016sfm, schoenberger2016mvs}, which follows an incremental procedure: detecting and matching local features, estimating relative poses through epipolar geometry, triangulating 3D points, and jointly refining camera parameters and structure via bundle adjustment. Although this purely geometric formulation exhibits inherent scale ambiguity and relies heavily on robust feature correspondences, it remains a well established framework. OpenSfM~\cite{adorjan2016opensfm} provides a similar classical approach.
        
        Recent deep learning methods have also been proposed for camera pose estimation and monocular depth prediction~\cite{zhou2017unsupervised, mihalea2023leveraging, yang2024depth, leroy2024grounding, keetha2025mapanything}. However, many of these approaches depend on training data generated using COLMAP~\cite{reizenstein2021common, ling2024dl3dv}, synthetic datasets~\cite{thabet2020self, deitke2023objaverse}, or datasets augmented with additional sensors~\cite{yeshwanth2023scannet++, xia2024rgbd}. While such methods often aim to recover metric scale, this is not required in our setting, as our approach is agnostic to absolute scale. Consequently, classical geometric pipelines offer a transparent and reliable means of maintaining scale consistency over long sequences, an aspect that can be more difficult to ensure with black-box neural network models.    

    \subsection{Ego-Trajectory Estimation using Image Segmentation}
        Segmentation-based trajectory estimation differs fundamentally from conventional drivable-area segmentation. While drivable-area segmentation identifies all regions where a vehicle could potentially drive, trajectory estimation predicts specific, dynamically feasible paths that the ego vehicle can realistically execute given scene context and motion constraints.
        
        Prior work on monocular odometry-based trajectory estimation typically follows other directions. Some jointly learn camera pose and monocular depth within unified networks~\cite{zhou2017unsupervised, mihalea2023leveraging}, while others rely on ground-truth data or additional sensors like LiDAR for supervision~\cite{li2018undeepvo, gupta2025diffusion}. Multi-modality has been addressed via multi-head architectures, where each head predicts a distinct motion mode~\cite{zhou2017unsupervised, mihalea2023leveraging}, or through diffusion-based models that iteratively sample diverse trajectory hypotheses~\cite{gupta2025diffusion}.
        
        Notably, \cite{barnes2017find} proposes a weakly supervised framework that decouples depth estimation from segmentation while leveraging LiDAR, demonstrating generalization across multiple travel modes in a single-head formulation. Following \cite{barnes2017find}, works like~\cite{ma2023self} refined the use of LiDAR, while \cite{sun2020see} developed a way to use GPS tracks. Our work shares this principle of separating geometric reasoning from segmentation-based prediction, but differs by not requiring LiDAR and emphasizing scalable, data-driven generalization across trajectory modes. Unlike multi-head or diffusion-based methods, our framework maintains architectural simplicity while enabling efficient multi-modal ego-trajectory estimation. Furthermore, because our pipeline relies exclusively on unconstrained monocular video, it naturally transfers to domains where acquiring strictly calibrated multi-sensor datasets is practically infeasible. This makes our approach uniquely suited for emerging autonomy applications such as off-road navigation and micro-mobility which generally lack the structured data infrastructure of traditional automotive environments.
        
\section{Methodology}
\label{sec:methodology}
    \subsection{Training Data Overview}
    The video-sharing platform YouTube hosts an estimated 20 billion videos, comprising over 3.9 billion hours of footage~\cite{youtube_blog_20_years, statista_youtube_format}, constituting the largest publicly accessible repository of video data in existence. Within this massive corpus, crowdsourced dashcam recordings represent an expansive and continuously growing source of real-world driving scenarios. While the exact volume of these specific recordings remains unquantified, conservative estimates suggest that the subset yields hundreds of thousands of hours of driving footage. Consequently, this platform offers an unprecedented, naturally diverse, and highly scalable data reservoir.

    Traditional autonomous driving datasets, such as KITTI~\cite{Kitti_dataset}, NuScenes~\cite{caesar2020nuscenes}, and Argoverse~\cite{argoverse} provide multiple modalities of highly structured, precisely calibrated, and meticulously labeled data. In stark contrast, crowdsourced dashcam videos consist exclusively of unlabeled, uncalibrated, and monocular footage. 
    
    We complied our dashcam dataset by processing a 30-hour subset of footage sourced from videos ranging from 20 minutes to 3 hours in duration. The raw video is spatially resized to a resolution of 720x1280 and temporally subsampled at 10 fps. Through our automated SfM pipeline, we successfully output over 650,000 annotated frames to serve as training data for the segmentation model.
    


    This data subset encompasses a diverse array of environmental conditions, including varying weather (e.g., dry, rain, snow), diurnal variations (morning, day, evening, night), and geographical representation across 14 distinct cities. Furthermore, the dataset captures significant hardware heterogeneity, featuring varied camera specifications (e.g., wide versus narrow field of view (FoV), varying lens distortion) and disparate mounting configurations ranging from low-profile car hoods to elevated truck windshields. Notably, our pipeline demonstrates robustness to suboptimal ``in-the-wild'' data artifacts, successfully handling videos plagued by watermarks, severe lens distortion, unorthodox camera orientations, and abrupt intra-video transitions.

\begin{figure}
    \centering
    \includegraphics[width=0.32\linewidth]{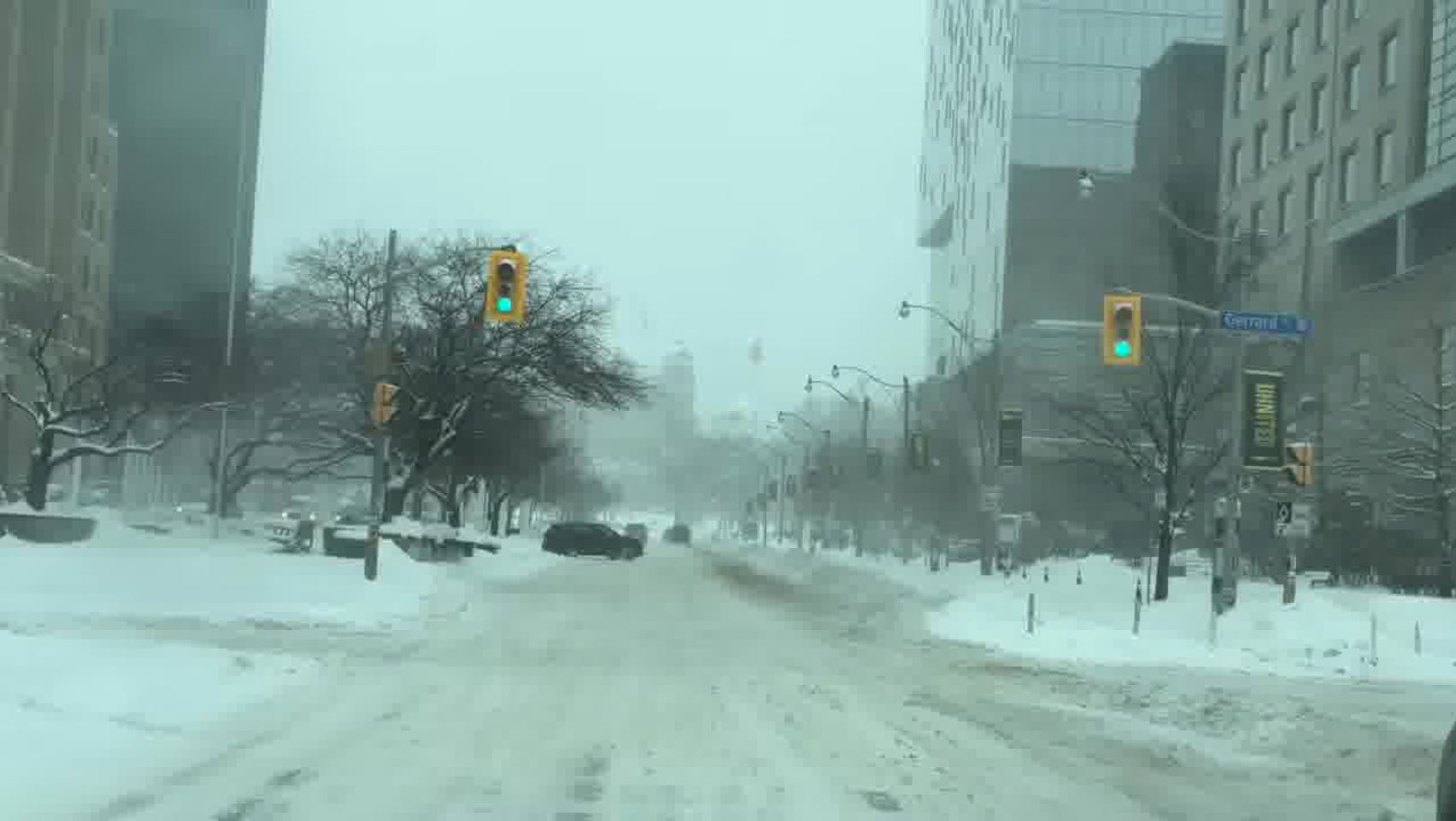}
    \vspace*{.5mm}
    \includegraphics[width=0.32\linewidth]{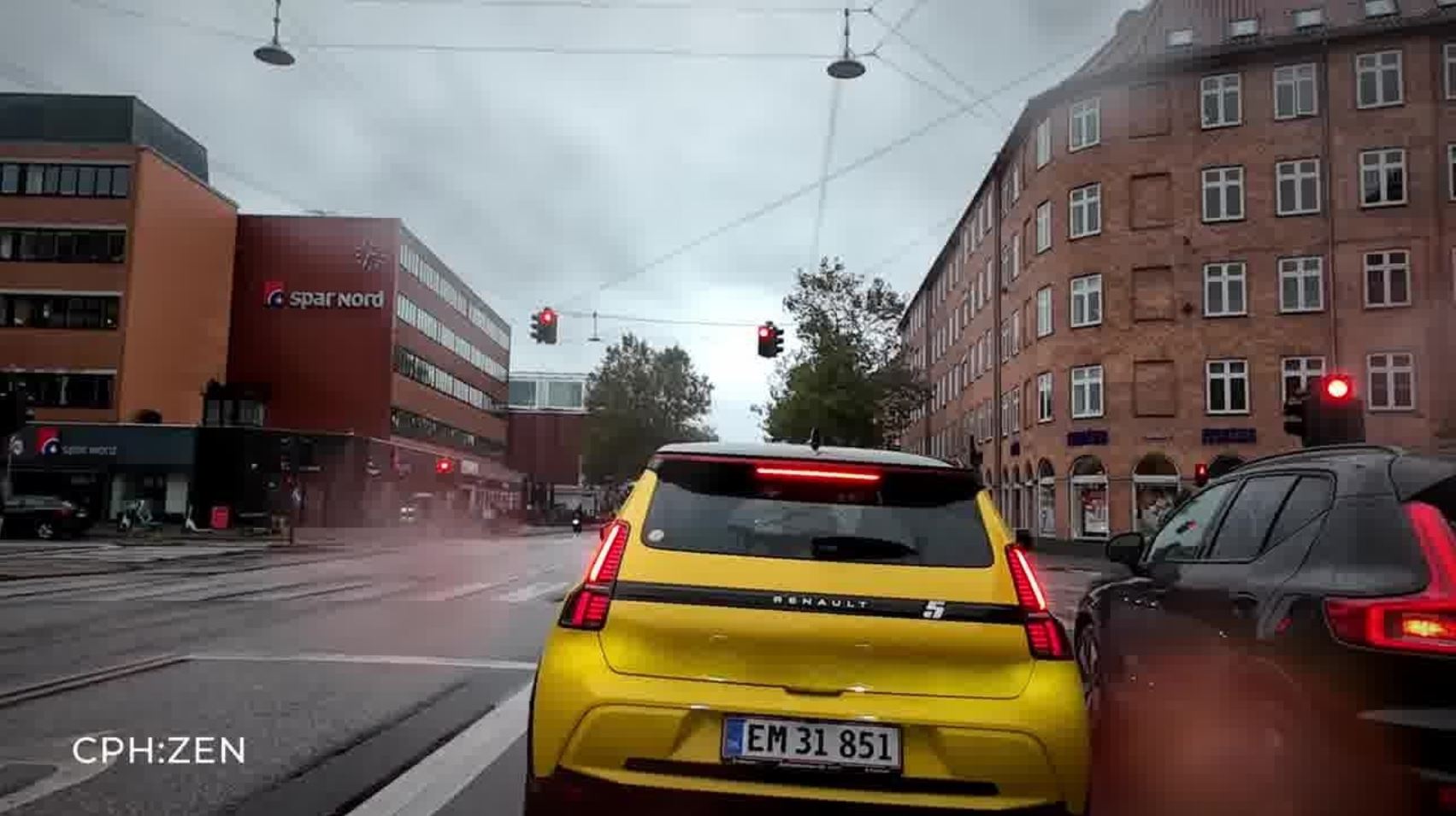}
    \vspace*{.5mm}
    \includegraphics[width=0.32\linewidth]{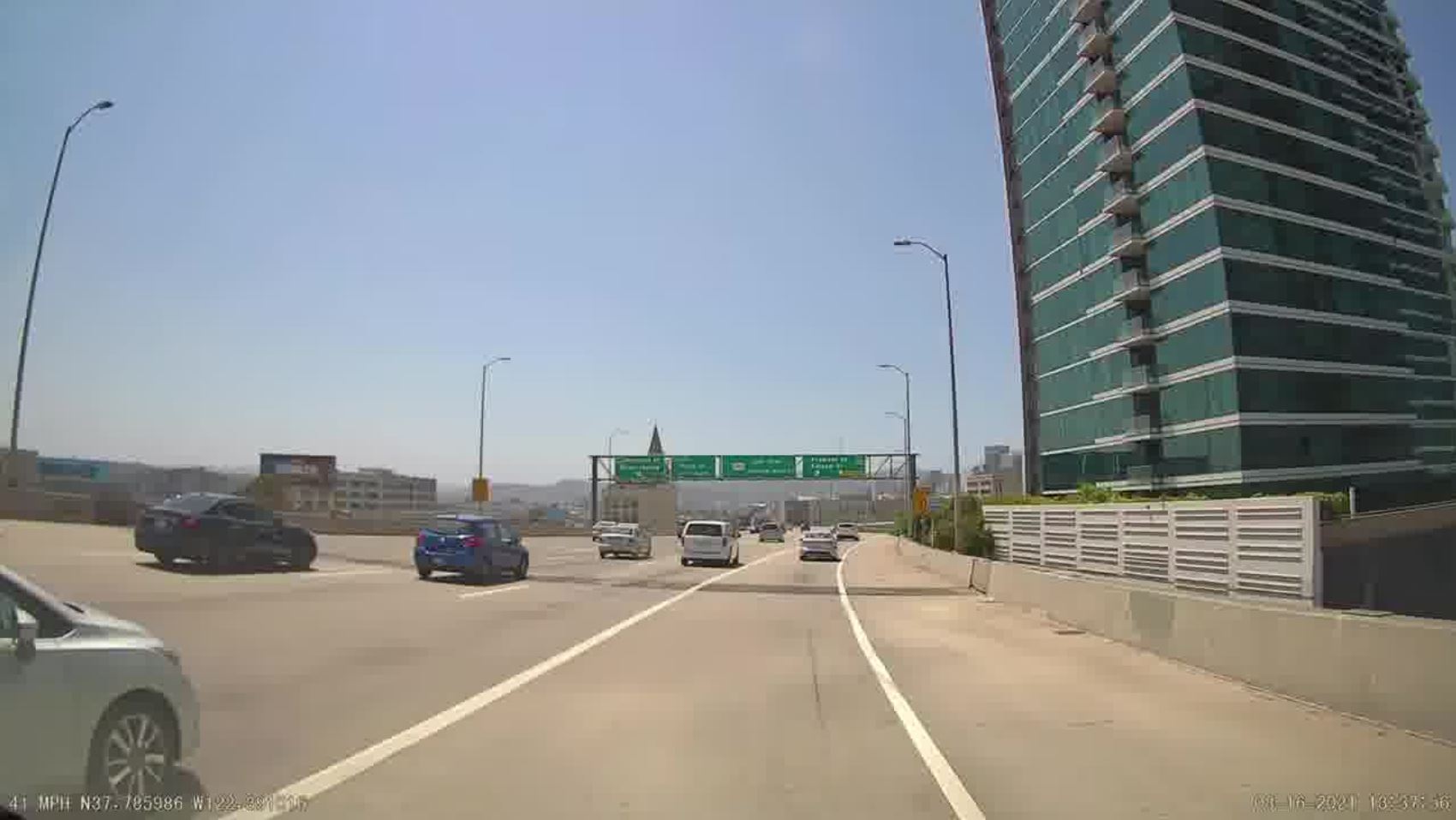}
    \vspace*{.5mm}
    \includegraphics[width=0.32\linewidth]{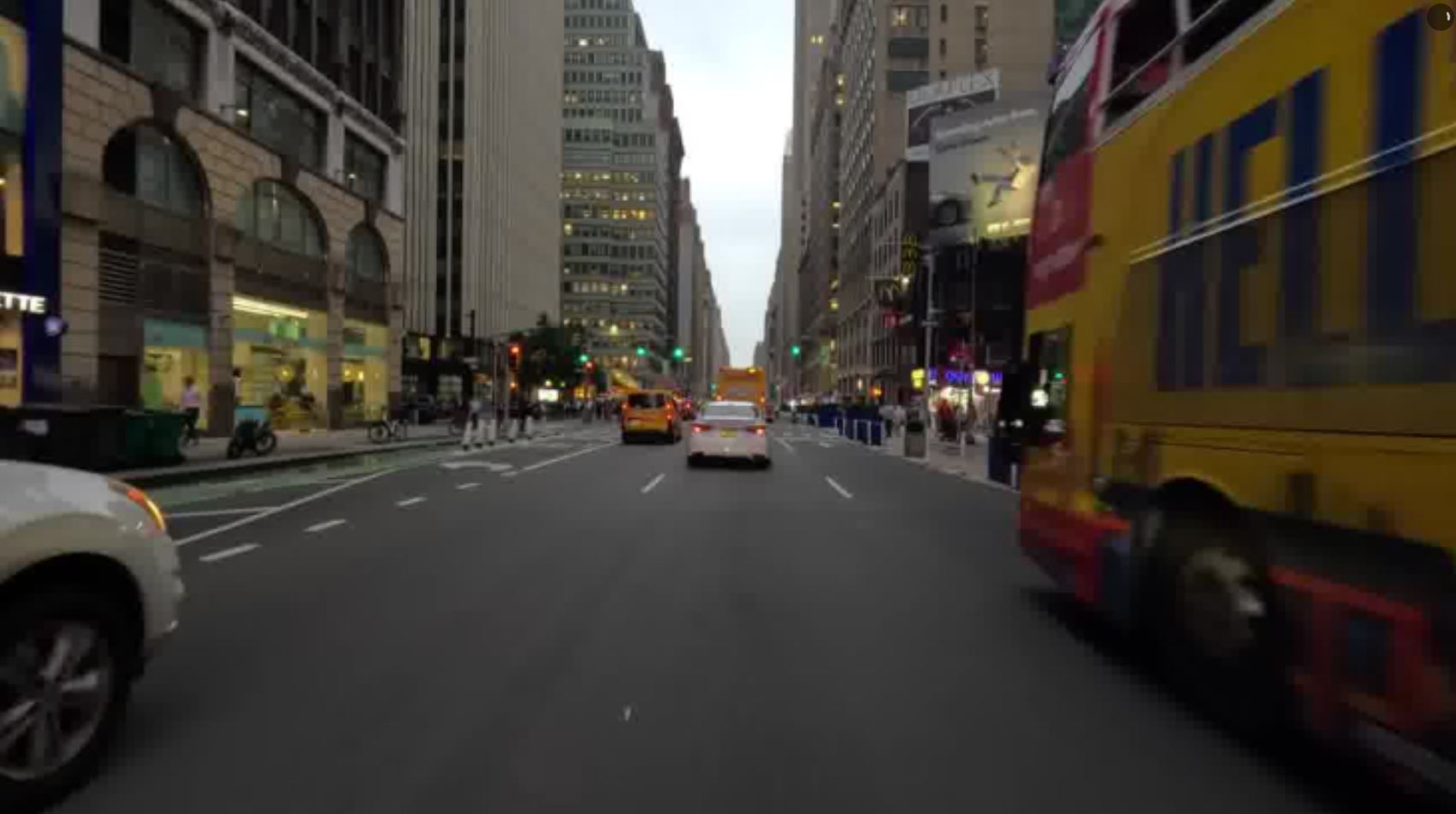}
    \includegraphics[width=0.32\linewidth]{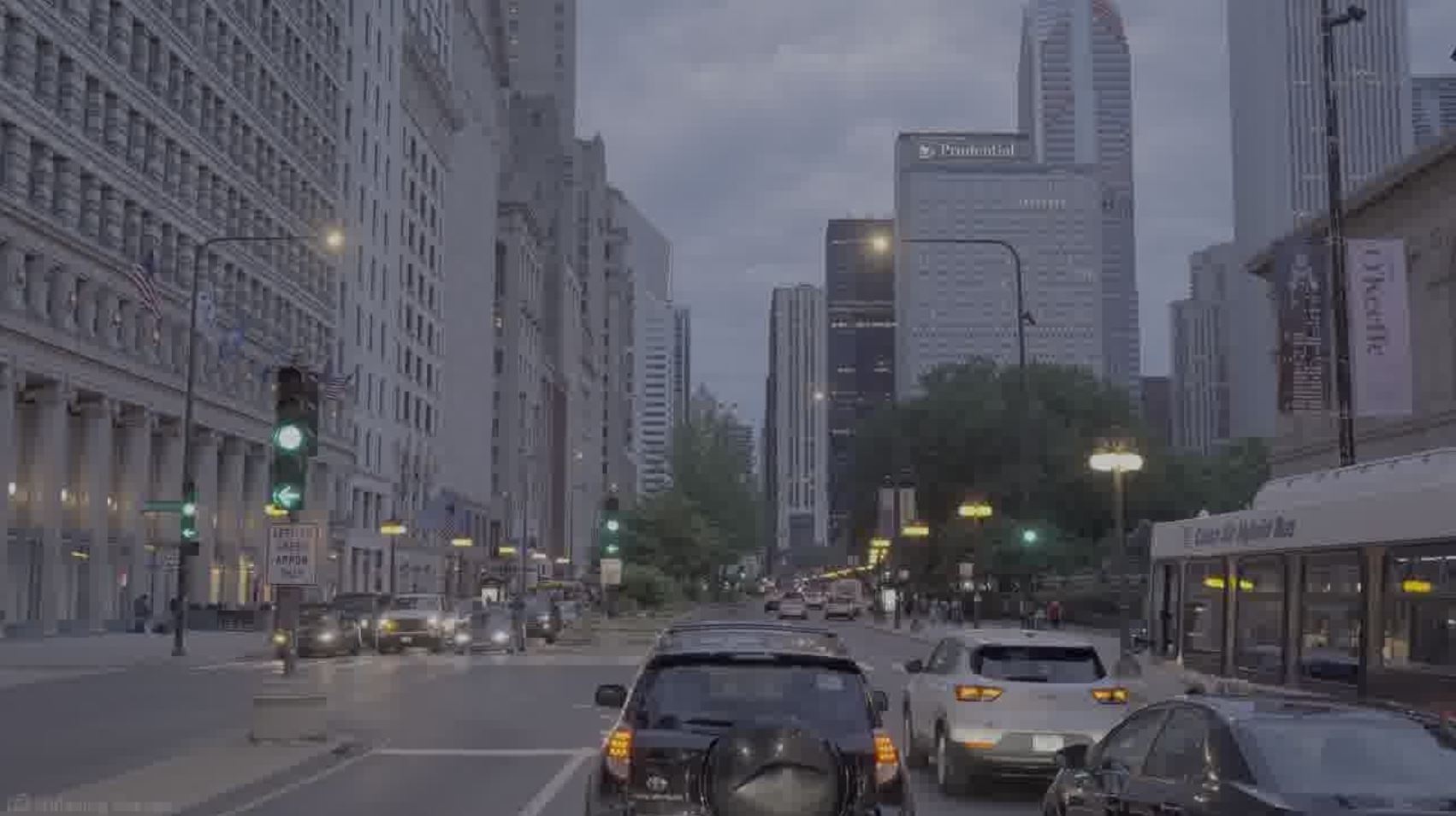}
    \includegraphics[width=0.32\linewidth]{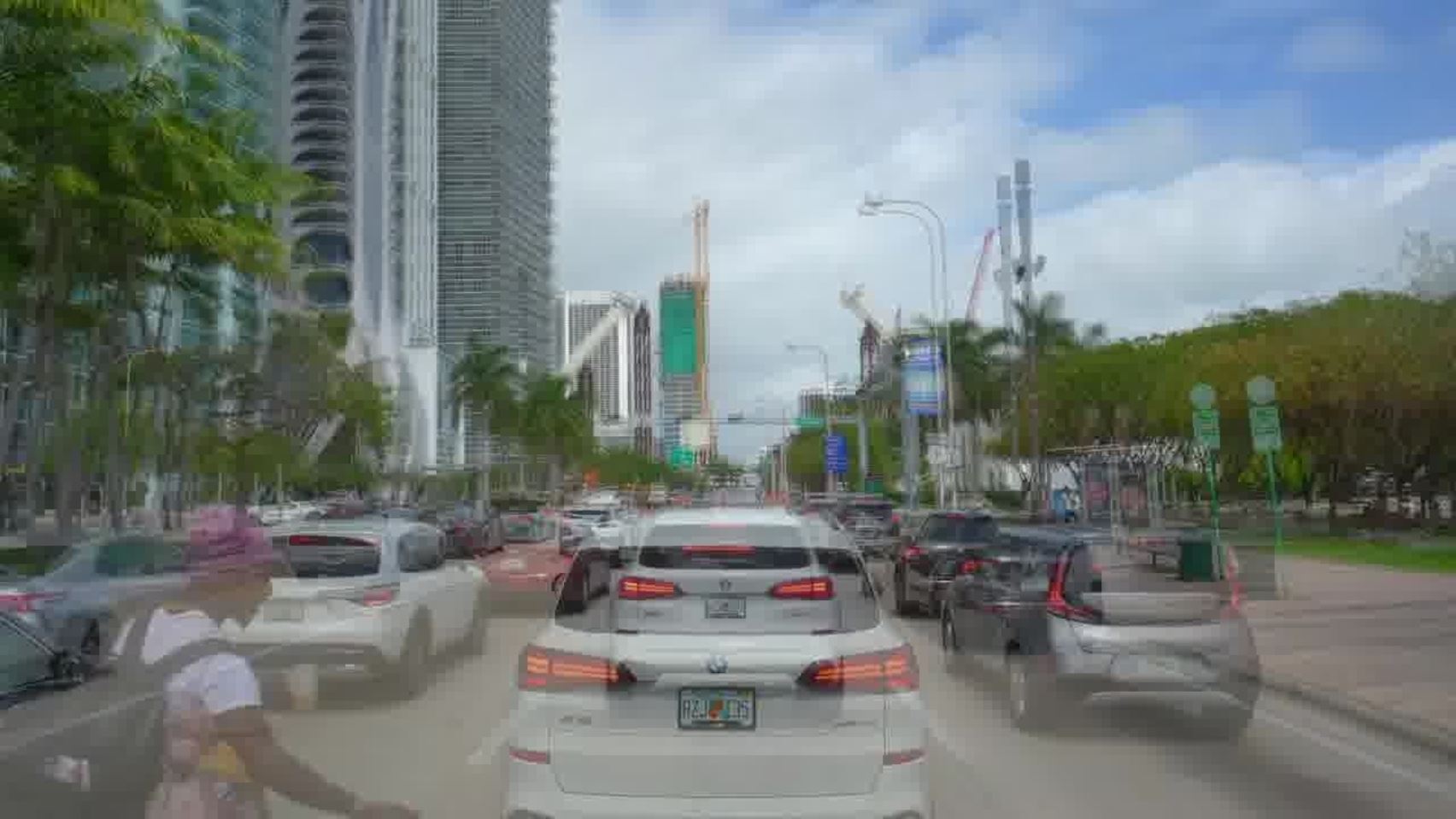}
    \caption{Qualitative examples demonstrating the diversity and unstructured nature of the collected dashcam dataset. (Top Left) Wide-angle FoV with moderate roll, captured in snowy conditions. (Top Center) Camera exhibiting an off-axis tilt, rain artifacts, and a visible watermark. (Top Right) A camera with wide-angle FoV and high barrel distortion. (Bottom Left) A very low mounted, narrow FoV camera with pin-cushion distortion. (Bottom Center) High-mounted camera perspective recorded during evening hours. (Bottom Right) Frame illustrating suboptimal in-the-wild artifacts, specifically an intra-video fade transition.}
    \label{fig:data_examples}
\end{figure}
    
    \subsection{Structure-from-Motion Pipeline}
    We employed the COLMAP~\cite{schoenberger2016sfm, schoenberger2016mvs} SfM pipeline, leveraging the Ceres Solver~\cite{Agarwal_Ceres_Solver_2022} for GPU-accelerated optimization, to process the video sequences. Specifically, we utilized SIFT feature extractor, sequential matcher, and incremental mapper modules to generate sparse 3D reconstructions. To promote robust, tightly coupled reconstructions, instead of expansive, loosely connected ones, we constrained the feature extraction to a maximum of 1,000 features per frame, and restricted the matcher to a maximum matching ratio of 0.7, and the minimum number of inliers to 30. We also assumed all inputted images follow the simple radial fisheye camera model, and that all images within a video are taken by the same camera. Because the time complexity of global bundle adjustment during the mapping phase scales at $O(N^4)$, processing the entire dataset sequentially is computationally intractable. To mitigate this limitation, we partitioned the video sequences into independent 20-minute (12,000-frame) intervals, which allowed for efficient, highly parallelized processing.  Ultimately, the sparse reconstruction yields 6-degrees-of-freedom (DoF) camera poses, estimated intrinsic camera parameters, and both local (per frame) and global 3D keypoint point clouds. These 6-DoF poses are parameterized as rigid world-to-camera transformations consisting of a unit quaternion $q = [q_w, q_x, q_y, q_z]^\top$ and a translation vector $\mathbf{t} \in \mathbb{R}^3$.

\subsection{Post Processing}\label{subsec:pp}
    To recover the ego-trajectory, the unscaled 6-DoF camera poses derived from the SfM pipeline must be spatially grounded. While SfM reconstructions are geometrically consistent, they are inherently ambiguous up to an unknown scale factor. Resolving this scale typically requires specialized sensor calibration or known reference distances within the scene. Given the unstructured nature of our video data, such explicit physical reference points are unavailable.

    To establish the scale and extract the precise height of the camera from the unscaled SfM reconstruction, we implemented a robust, projection-based ground plane estimation system. This procedure leverages the sparse 3D point cloud and camera poses generated by COLMAP, processing them across sequential frames to calculate a statistically robust global height estimate. Fundamentally, estimating the camera height reduces to finding the orthogonal distance between the camera's optical center and the localized ground plane in the reconstruction space. To help achieve this, we make two critical assumptions that hold throughout every video:
    
    \begin{itemize}
        \item \textbf{Kinematic Motion Model:} The vehicle's ego-motion adheres to a kinematic bicycle model, such that the camera is rigidly coupled to the vehicle and its trajectory directly represents the path of the physical car body.
        \item \textbf{Geometric Alignment and Stability:} The camera's optical axis ($z$-axis, by convention) is approximately parallel to the local tangent plane of the road surface. Furthermore, the vertical displacement between the camera's optical center and the ground plane is assumed to be invariant throughout each reconstruction.
    \end{itemize}
    
    We achieve the estimation by isolating 3D keypoints that heuristically correspond to the road surface. Assuming the local road surface is essentially planar, we project these points into the longitudinal depth-height plane by marginalizing the lateral $x$-axis (defined conventionally as the left-right axis in the ego-vehicle frame). We then apply a Theil-Sen estimator to the remaining depth and height coordinates. The resulting line of best fit represents the longitudinal profile of the ground plane. Because the local coordinate system originates at the camera's optical center, the $y$-intercept of this robustly fitted line directly yields the unscaled height of the camera above the ground.
    
    \subsubsection*{\textbf{Height Estimation}}

    The height estimation pipeline is structured into two distinct phases: first, the construction of a comprehensive global ground set; second, a localized robust regression to determine the precise height of the camera. 
    
    In the first phase, we systematically identify and accumulate all valid ground points across the entire sequence. For a given camera $i$, we evaluate its visible 3D points, $\mathbf{p}_w \in \mathbb{R}^3$, in the global SfM coordinate frame. To ensure physical observability and establish the point's relative position, these points are rigidly transformed into the camera's local coordinate frame: $\mathbf{p}_c = \mathbf{R}_{cw, i} \mathbf{p}_w + \mathbf{t}_i$
    where $\mathbf{R}_{cw, i}$ and $\mathbf{t}_i$ denote the rotation matrix and translation vector for camera $i$, respectively, and $\mathbf{p}_c = [x_c, y_c, z_c]^\top$. The observable points are projected onto the 2D image plane and filtered against a heuristic region of interest, $\mathcal{R}$, which bounds the lower-central sector of the image where the road surface is expected to appear: $\mathcal{R} = \{ (u, v) \mid 0.20 W < u < 0.80 W, \; v > 0.75 H \}$ where $W$ and $H$ are the image width and height. Any 3D point that successfully projects into $\mathcal{R}$ is persistently appended to a comprehensive global ground set, $\mathcal{G}$. This cross-frame accumulation ensures a dense and reliable mapping of the ground surface, even if individual frames suffer from sparse feature extraction.
    
    In the second phase, once $\mathcal{G}$ is fully constructed, we iterate back through the sequence to estimate the height for each camera $i$. To mitigate computational overhead and enforce the assumption of local planarity, we query the fully populated $\mathcal{G}$ to extract only the localized subset of ground points situated within a maximum Euclidean distance threshold, $d_{max} = 1.0$ SfM units, from the camera's global optical center, $\mathbf{c}_{w, i} = -\mathbf{R}_{cw, i}^\top \mathbf{t}_i$. (Note that COLMAP reconstructions operate in an arbitrary, unscaled space where spatial extents are intrinsically small; anecdotally, one physical meter corresponds to $10^{-2}$ to $10^{-3}$ SfM units).
    
    This localized, distance-filtered subset of ground points is then transformed into the local coordinate frame of camera $i$. Extracting the local vertical coordinates $y_c$ and depth coordinates $z_c$ of this subset, we model the ground plane's longitudinal profile. To robustly estimate the linear relationship despite outliers caused by dynamic objects or irregular terrain, we apply a Theil-Sen estimator~\cite{theil1950rank,sen1968estimates}. It is a median-based non-parametric regression method that fits a line to the local $(z_c, y_c)$ pairs. The $y$-intercept of this fitted line, $y_{int}^{(i)}$, represents the robust estimate of the camera's height above the local ground plane. We ensure global consistency by calculating the median of all valid frame intercepts:
    \begin{equation}
        H_{cam} = \mathrm{median}( \{ y_{int}^{(1)}, y_{int}^{(2)}, \dots, y_{int}^{(N)} \} )
    \end{equation}

    \begin{figure}
        \centering
        \includegraphics[width=0.55\linewidth]{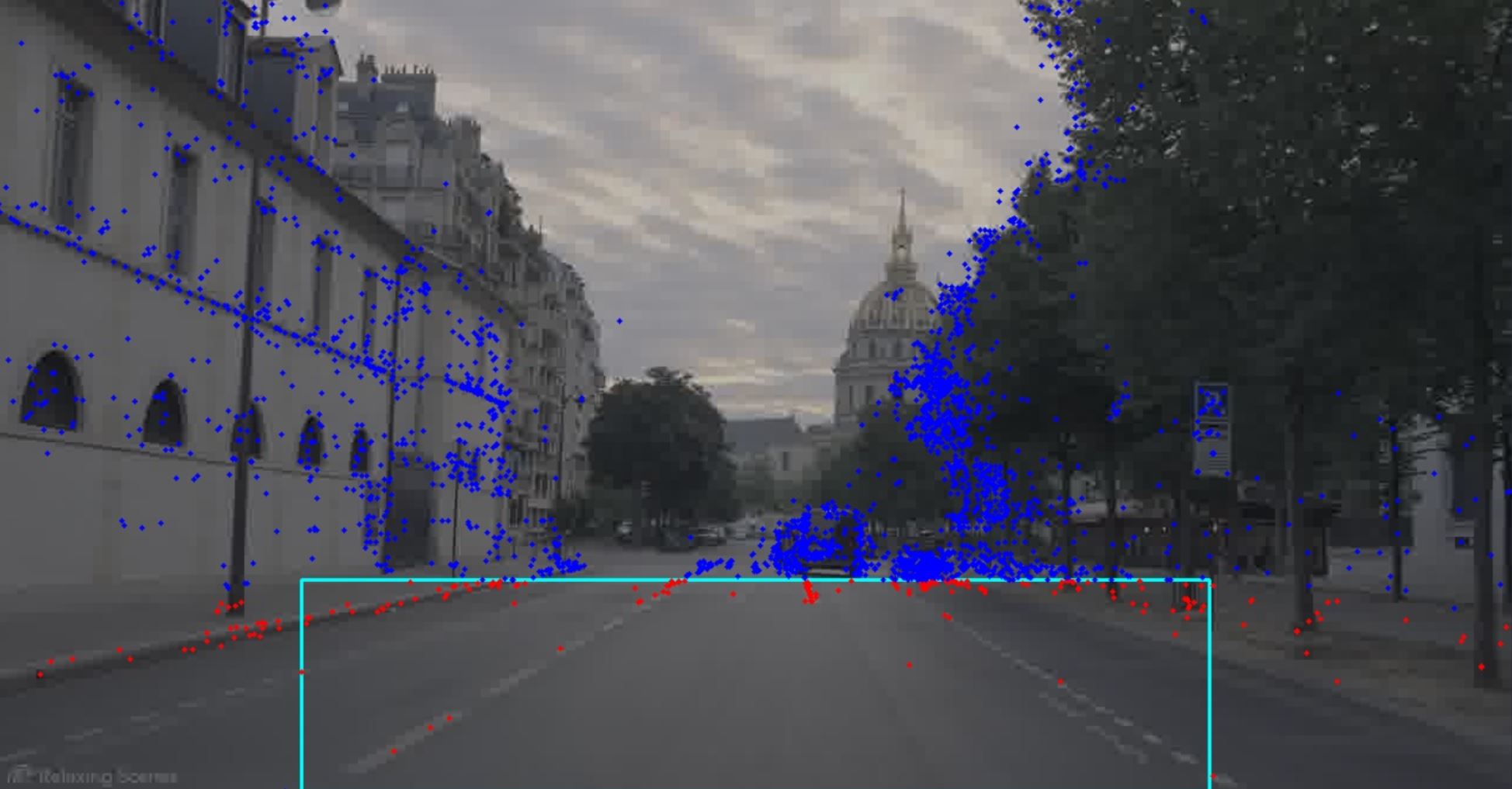}
        \includegraphics[width=0.35\linewidth]{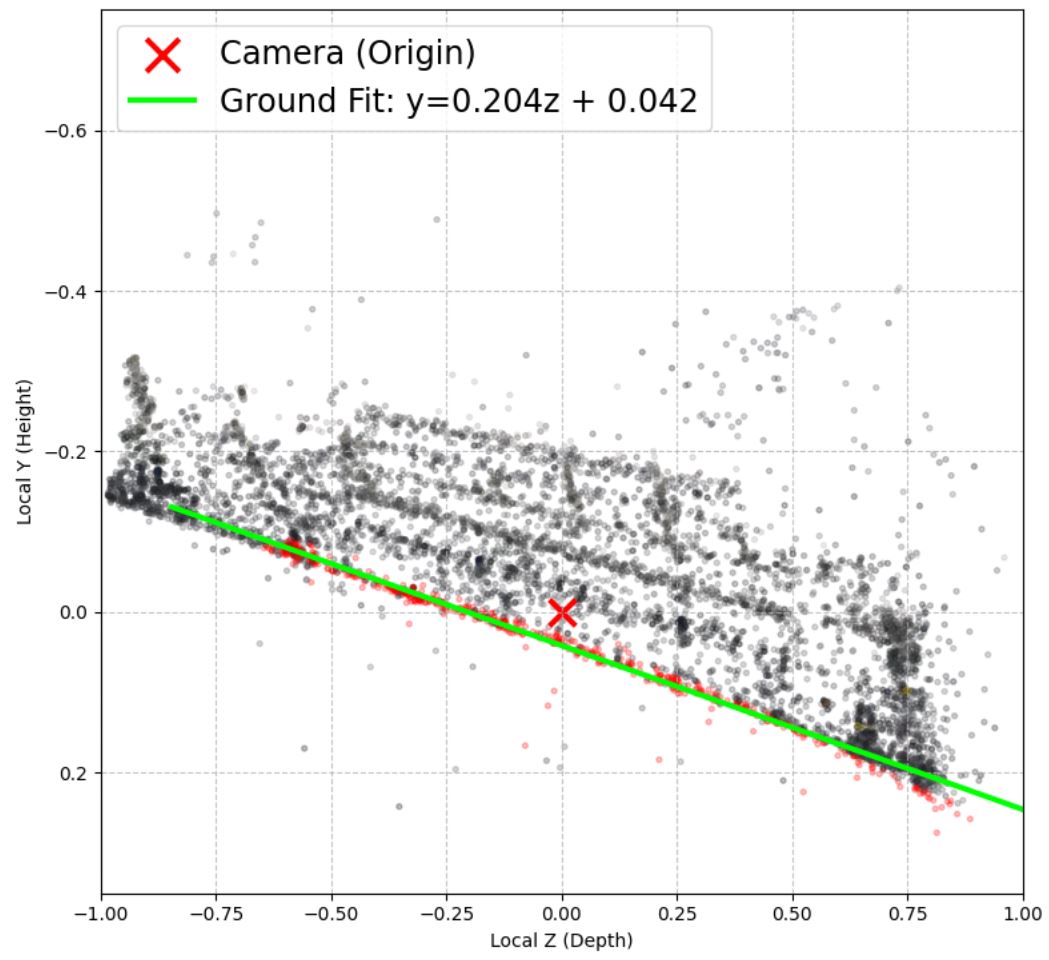}
        \caption{Visualization of the two-phase camera height estimation pipeline. (Left) Phase 1: Global ground set accumulation in progress. Red points denote 3D keypoints registered to the global ground set ($\mathcal{G}$). (Right) Phase 2: Localized height regression. A robust Theil-Sen estimator is applied to the spatial subset of ground points to compute the scale vertical offset.}
        \label{fig:height_calc}
    \end{figure}

    \subsubsection*{\textbf{Mask Generation}}

    Once the absolute camera height $H_{cam}$ is determined, this metric grounding is extended to derive a projected trajectory mask representing the ego-vehicle's path. Within the SfM coordinate space, we consider a temporal window of the subsequent 50 frames (5 seconds) of 3D camera coordinates. These coordinates are translated along the gravity vector by the calculated height: $$P_{ground} = P_{cam} + H_{cam} \mathbf{e}_y$$ where $\mathbf{e}_y$ denotes the unit vector directed downward within the camera's local reference frame. This transformation yields a grounded trajectory line representing the vehicle's footprint on the road surface.
    
    To generate a planar mask from this trajectory, a lateral width is assigned proportional to the camera height, specifically $1.5 \times H_{cam}$, to account for the vehicle's physical dimensions. This spatially grounded mask serves as the projected ego-motion trajectory and is utilized as the ground-truth input for supervising the segmentation model.

     Occasionally, COLMAP generates inaccurate trajectory reconstructions, where, despite ego vehicle movement, no (or an incorrect) trajectory is calculated. We identify ``null frames" where our pipeline detects inter-frame displacement in the reconstruction space exceeding $10^{-4}$ SfM units but fails to project a valid trajectory mask. This criterion effectively distinguishes between legitimate stationary periods (e.g., traffic signals) and erroneous motion estimates. Through this automated quality control, approximately 100,000 frames were discarded, resulting in the aforementioned high-fidelity 650,000 frame dataset. 

    \begin{figure}
        \centering
        \includegraphics[width=0.49\linewidth]{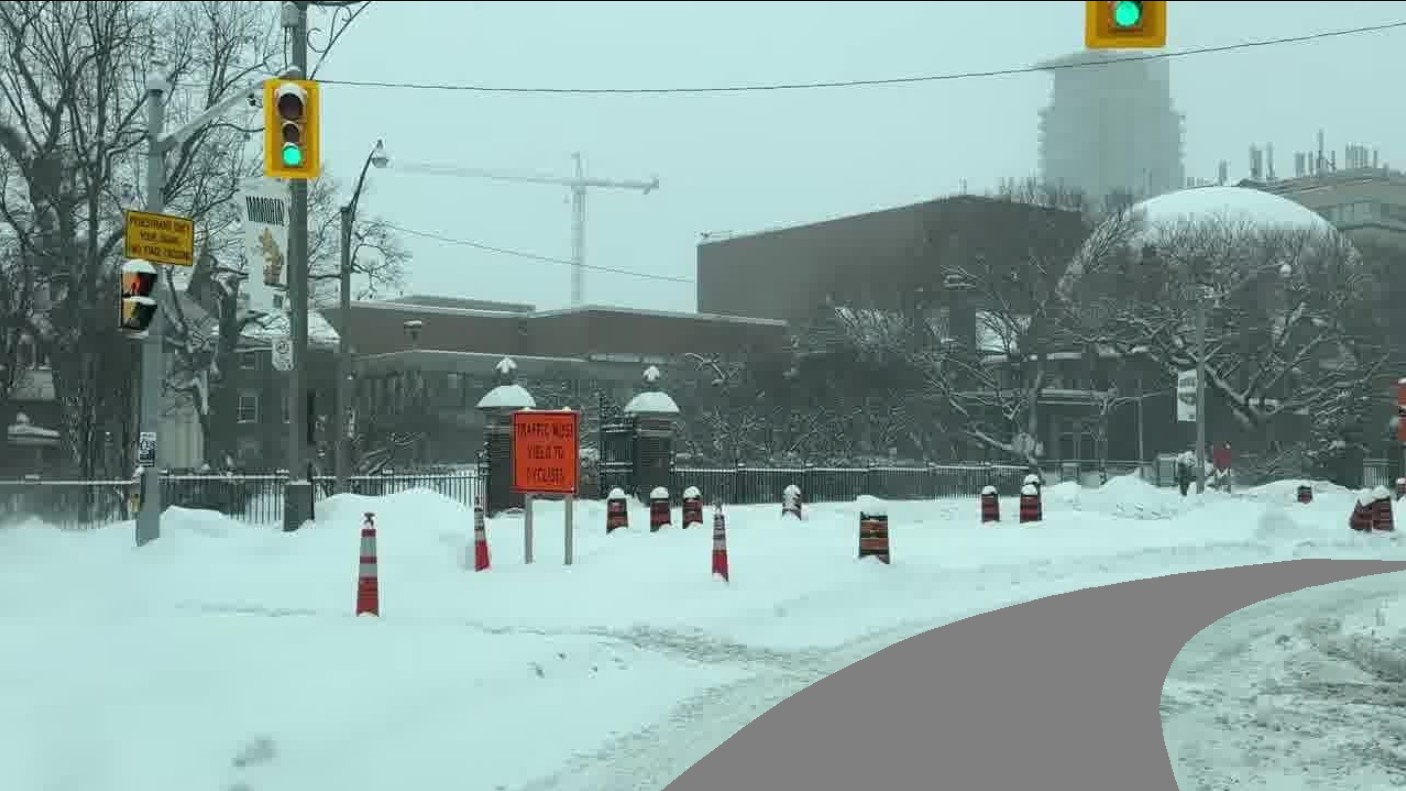}
        \vspace*{1mm}
        \includegraphics[width=0.49\linewidth]{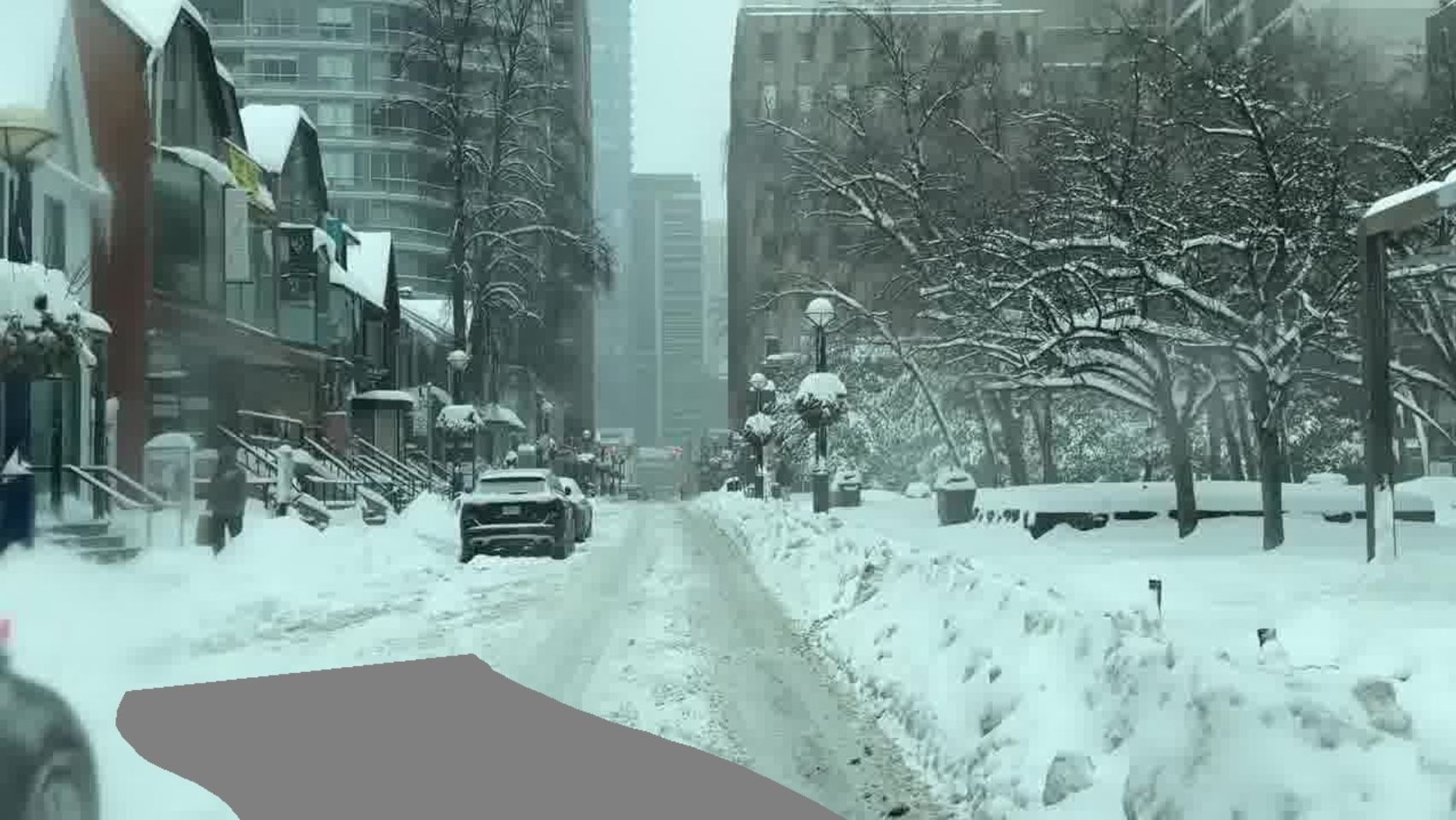}
        
        \includegraphics[width=0.49\linewidth]{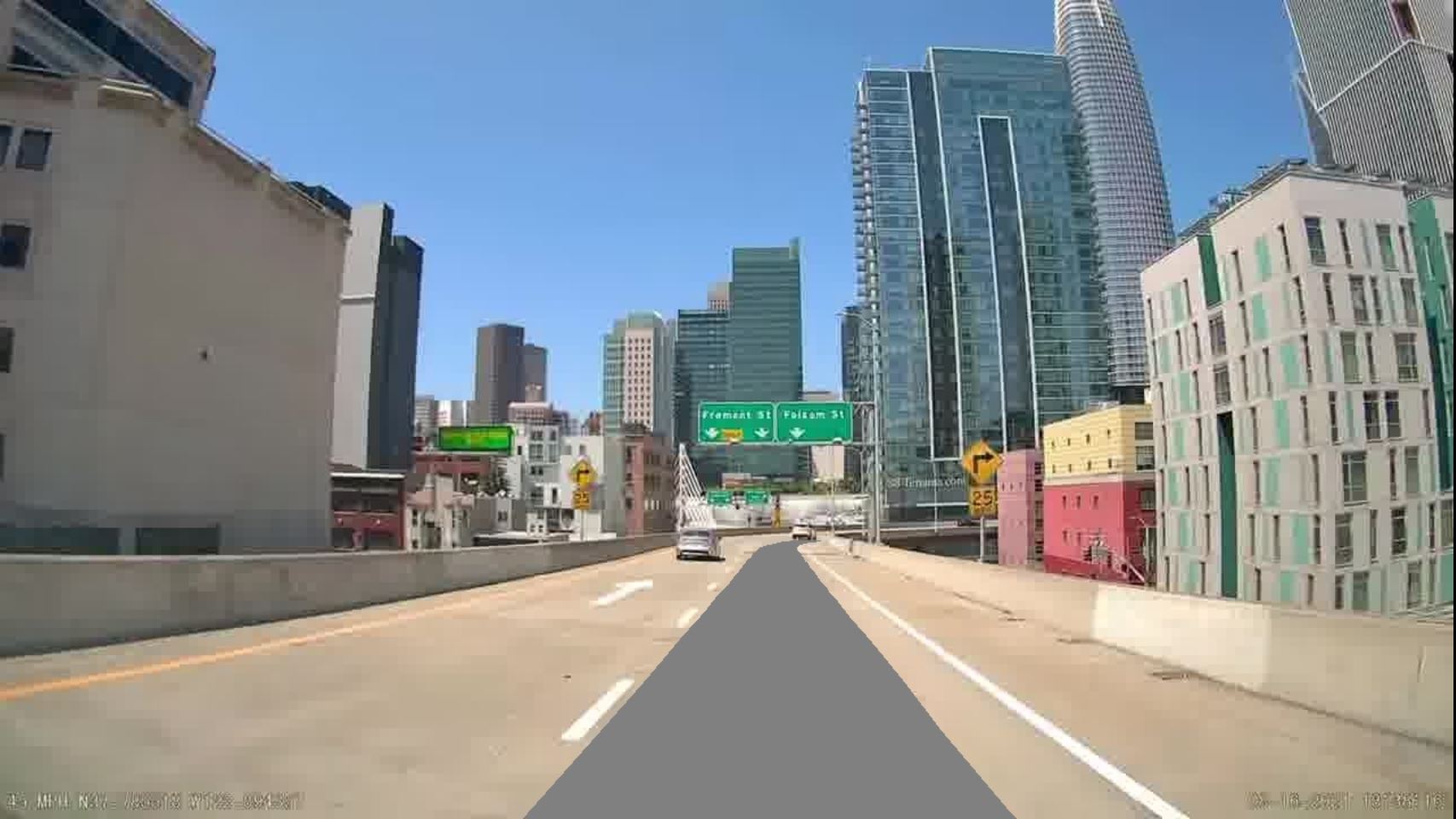}
        \vspace*{1mm}
        \includegraphics[width=0.49\linewidth]{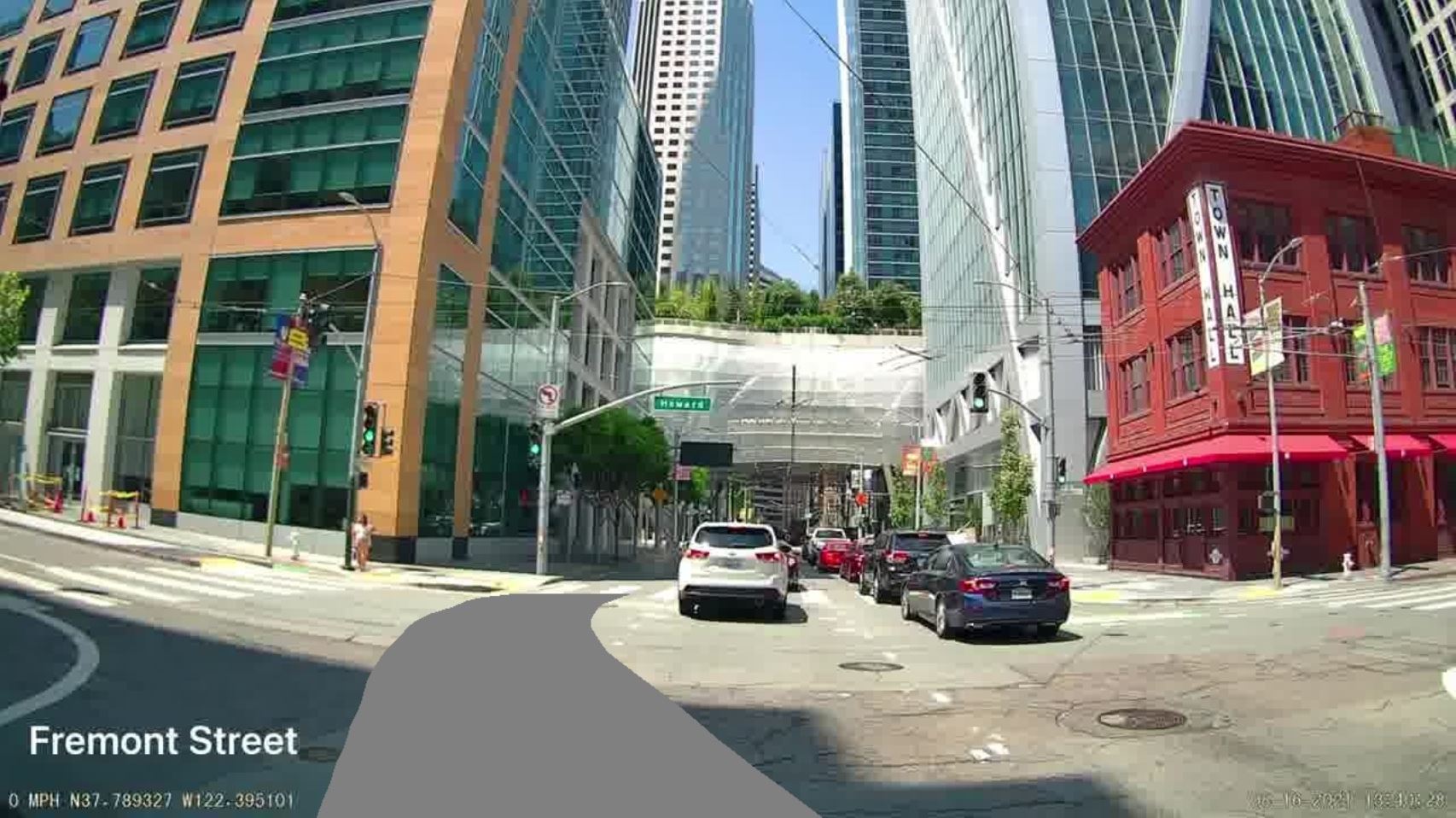}
        \caption{Examples of ground-truth masks generated via our pipeline, illustrating the ego vehicle: (Top Left) executing a sharp right turn, (Top Right) initiating a parallel parking maneuver, (Bottom Left) cruising along a straight highway, and (Bottom Right) performing a lane change.}
        \label{fig:gt_masks}
    \end{figure}



    \label{subsec:sfm-plane}

\subsection{Trajectory as Segmentation}
We model trajectory estimation as a binary image segmentation problem. Given a single RGB image, the network predicts a mask that represents feasible future ego motion on the ground plane.

We use a standard UNet architecture with a ResNet34~\cite{he2016deep} encoder backbone. The encoder is initialized with ImageNet~\cite{deng2009imagenet} pretrained weights. This provides strong low-level and mid-level visual features such as edges, textures, and scene structure. The decoder upsamples these features to produce a full resolution prediction. The final layer outputs a single-channel mask with raw logits. The supervision masks are generated from the projected ego trajectory.

While the supervision masks generated from projected ego-trajectories explicitly encode the single motion executed by the vehicle, our goal is to predict a broader set of feasible trajectory hypotheses from a single RGB input. Real-world driving environments naturally present multiple valid navigational choices. The model is therefore tasked with capturing this underlying structural prior, learning to infer the full distribution of valid paths from a dataset where only one such path is demonstrated per scene. To achieve this, we have designed a weighted negative log likelihood loss applied to the logits. Let $x$ denote the predicted logit and $y \in \{0,1\}$ denote the ground truth mask. 
The loss is defined as:

\begin{equation}
\label{eq:loss}
\mathcal{L}(x, y) = - (y - \epsilon)\,\log \sigma(x) + C
\end{equation}

where $\sigma(x)$ is the sigmoid function, $\epsilon$ is a small constant, and $C$ is a constant offset to ensure that the loss is non-negative.

This formulation reduces the penalty for false positives and increases the penalty for false negatives. In practice, this means that missing a feasible trajectory region is much more costly than predicting an extra region. As a result, the model is encouraged to cover all plausible motion regions rather than simply copying the single demonstrated path. If false positives were penalized more heavily, the model would collapse to predicting a single hypothesis that resembles the ground truth masks.

The log probability is clipped to prevent extremely large negative values. Without clipping, a very wrong prediction on a positive pixel would create an extremely large loss and unstable gradients. Clipping limits this explosion and stabilizes training. At the same time, the loss is asymmetric because it does not include the $\log(1 - \sigma(z))$ term. Background pixels are allowed to push logits strongly negative with little resistance. Positive pixels are pushed strongly toward large positive logits because false negatives are heavily penalized. As a result, logits become polarized. Feasible regions move toward strong positive values, and non-feasible regions move toward strong negative values. This polarization reduces the uncertain mask predictions and produces sharper, more structured masks.

To generate output masks, we use the probabilistic sigmoid output directly rather than thresholding it to a binary mask. This preserves confidence information and captures uncertainty near boundaries, which helps represent multiple plausible regions rather than forcing a single hard decision. As a result, the masks reflect the model's belief distribution over the drivable area.
\section{Experiments \& Results}
\begin{figure*}[!t]
    \centering

    \subfloat[Straight]{\includegraphics[width=0.24\linewidth]{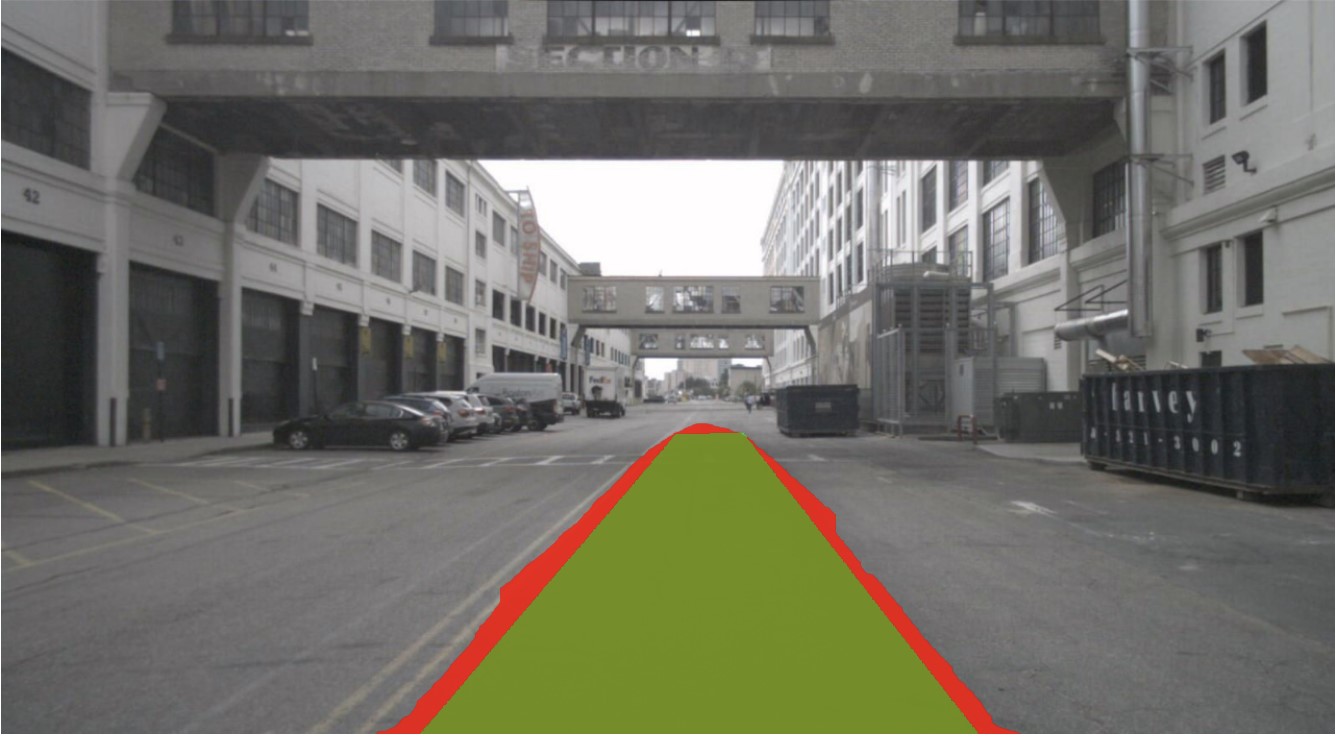}\label{fig:pretrain:Straight}}%
    \hfill
    \subfloat[Rainy]{\includegraphics[width=0.24\linewidth]{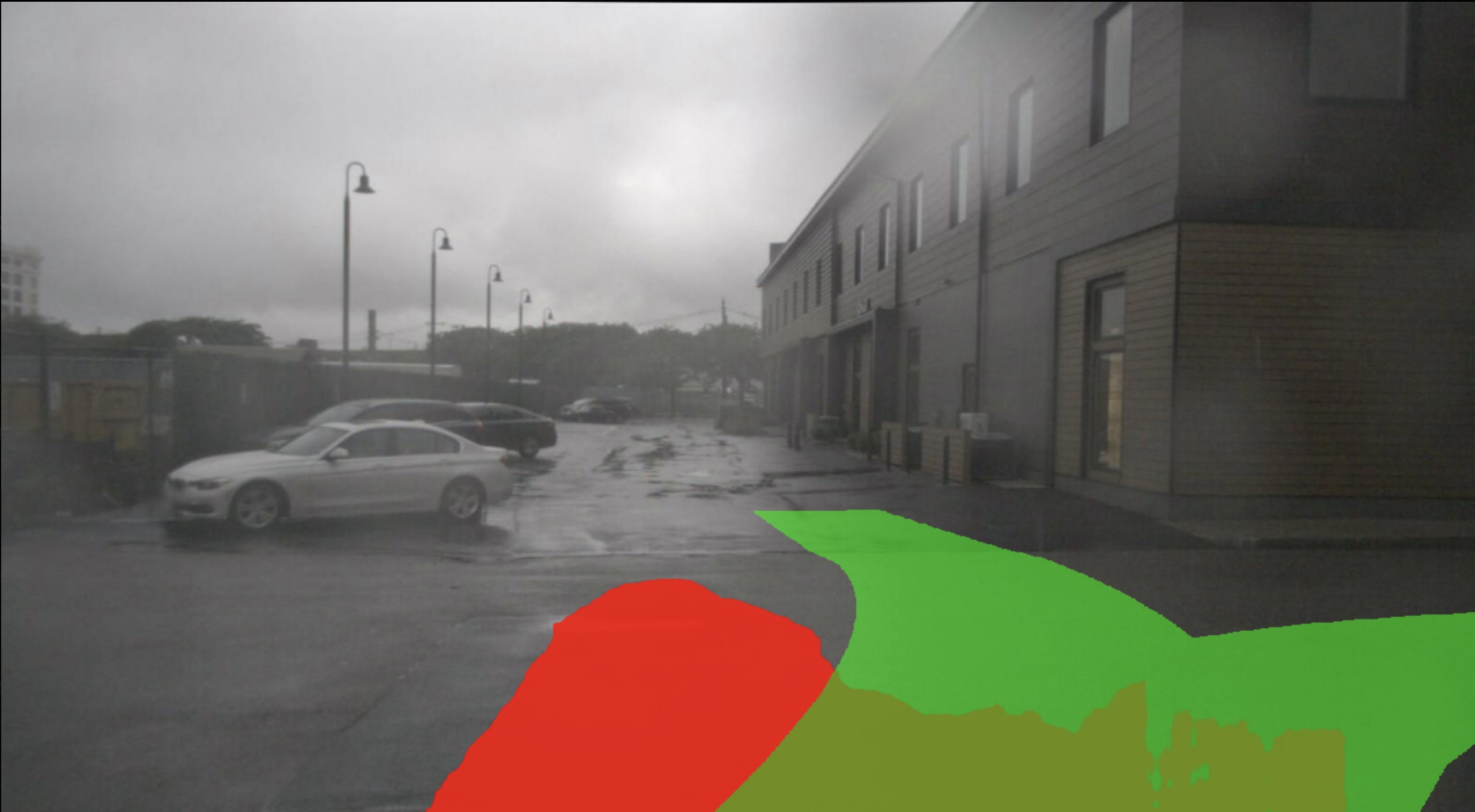}\label{fig:pretrain:rainy}}%
    \hfill
    \subfloat[Multiple Hypotheses]{\includegraphics[width=0.24\linewidth]{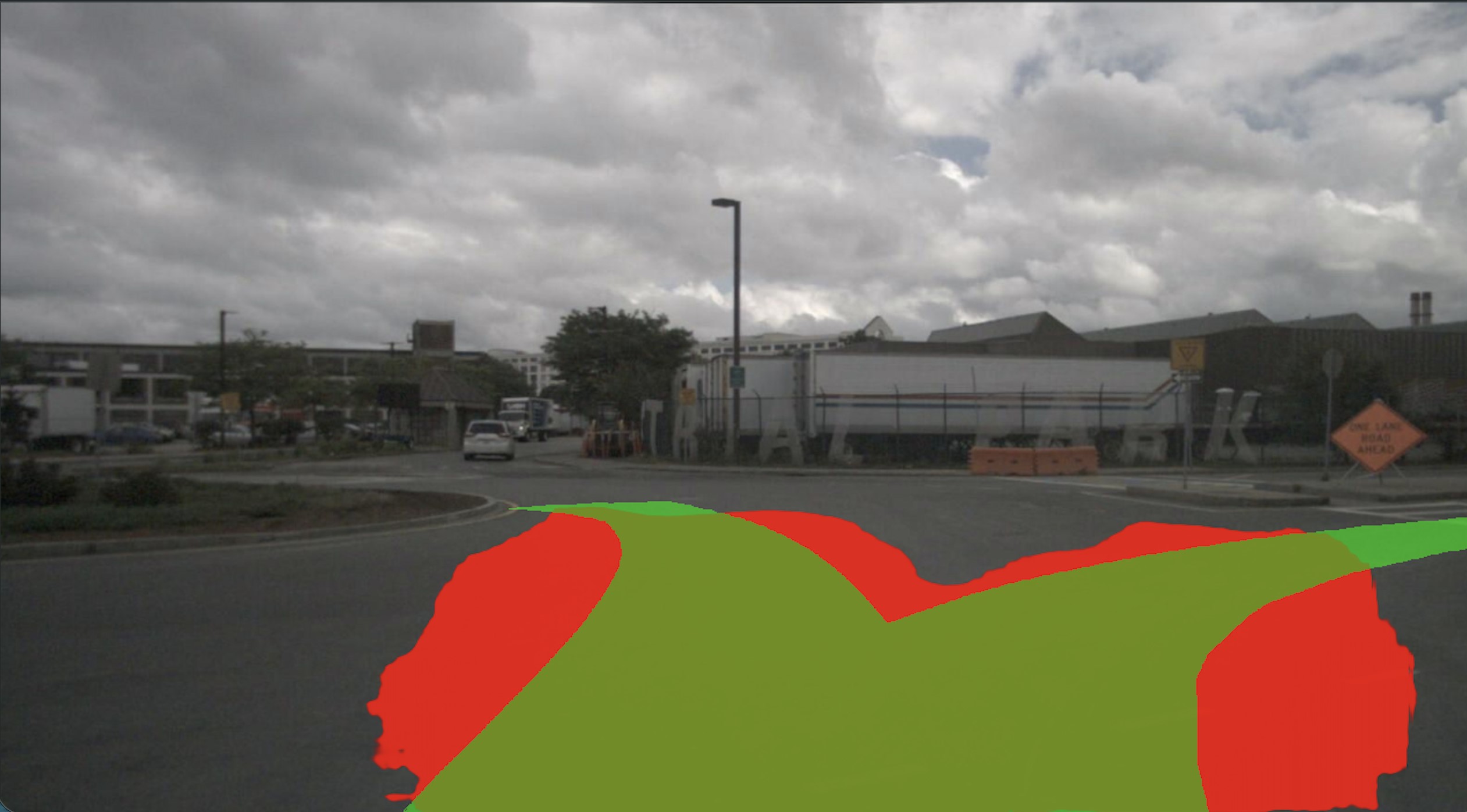}\label{fig:pretrain:multimode_gt}}%
    \hfill
    \subfloat[Road Intersection]{\includegraphics[width=0.24\linewidth]{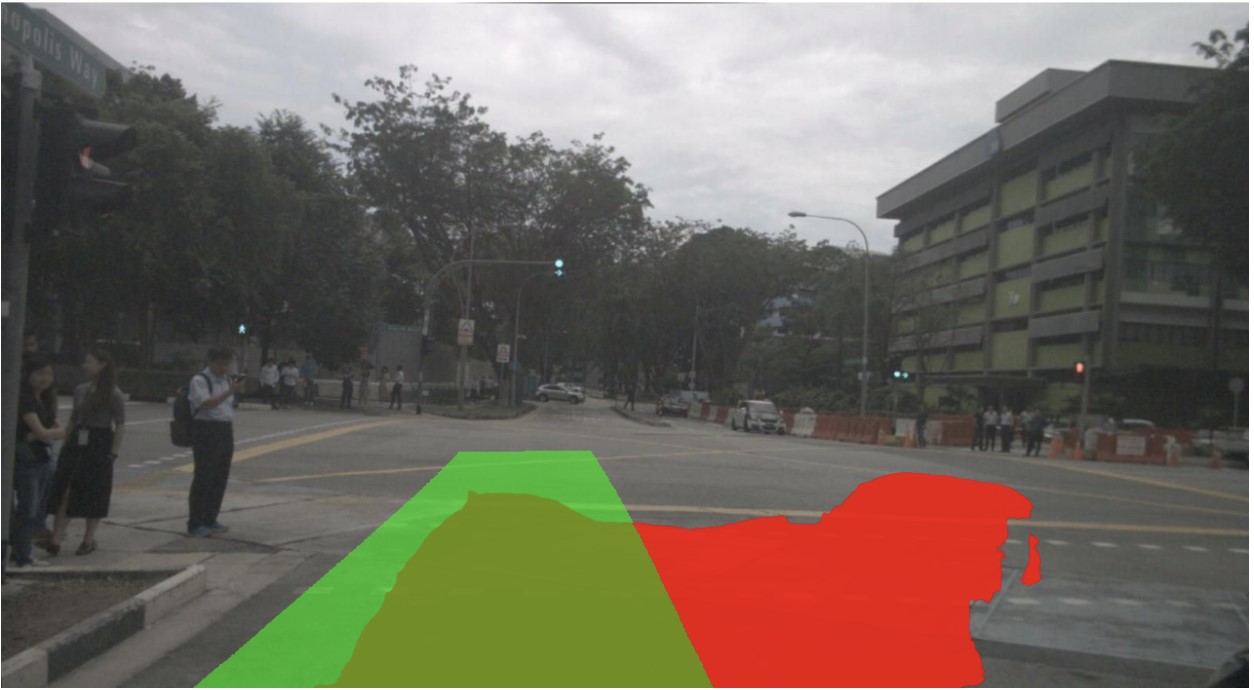}\label{fig:pretrain:Singapore_Multimode}}%

    \vspace{2pt}

    \subfloat[Road Split]{\includegraphics[width=0.24\linewidth]{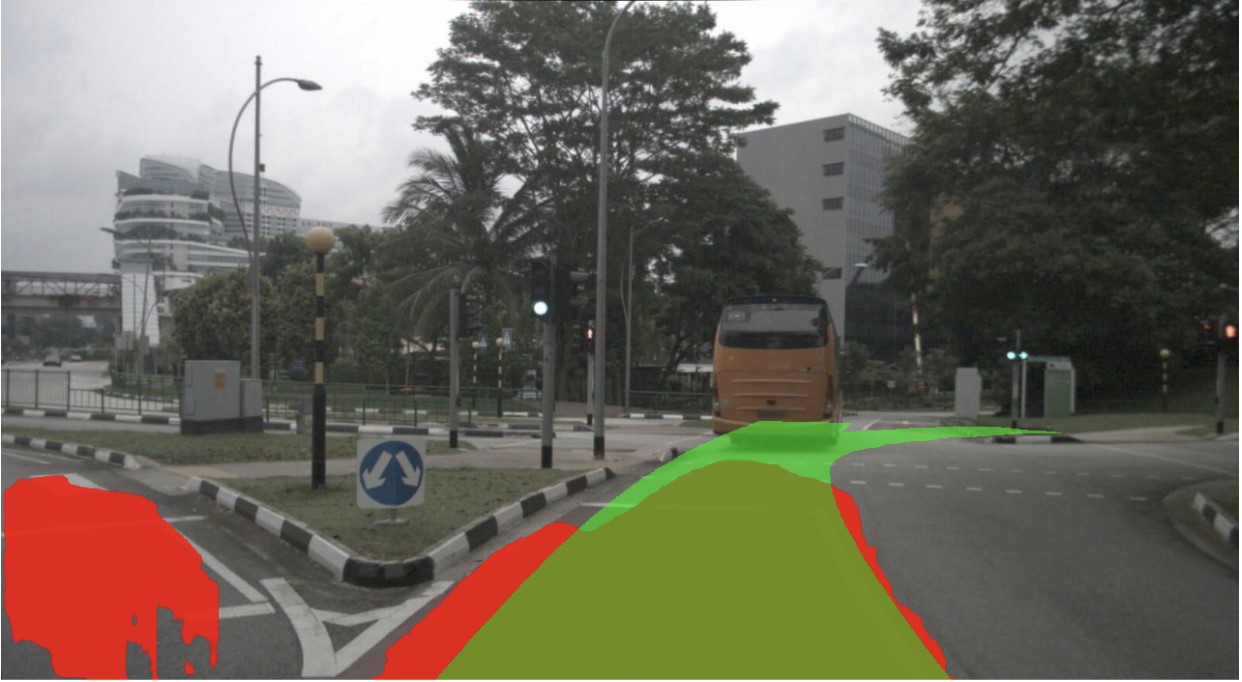}\label{fig:pretrain:Singapore_Diversion}}%
    \hfill
    \subfloat[Impossible Turn]{\includegraphics[width=0.24\linewidth]{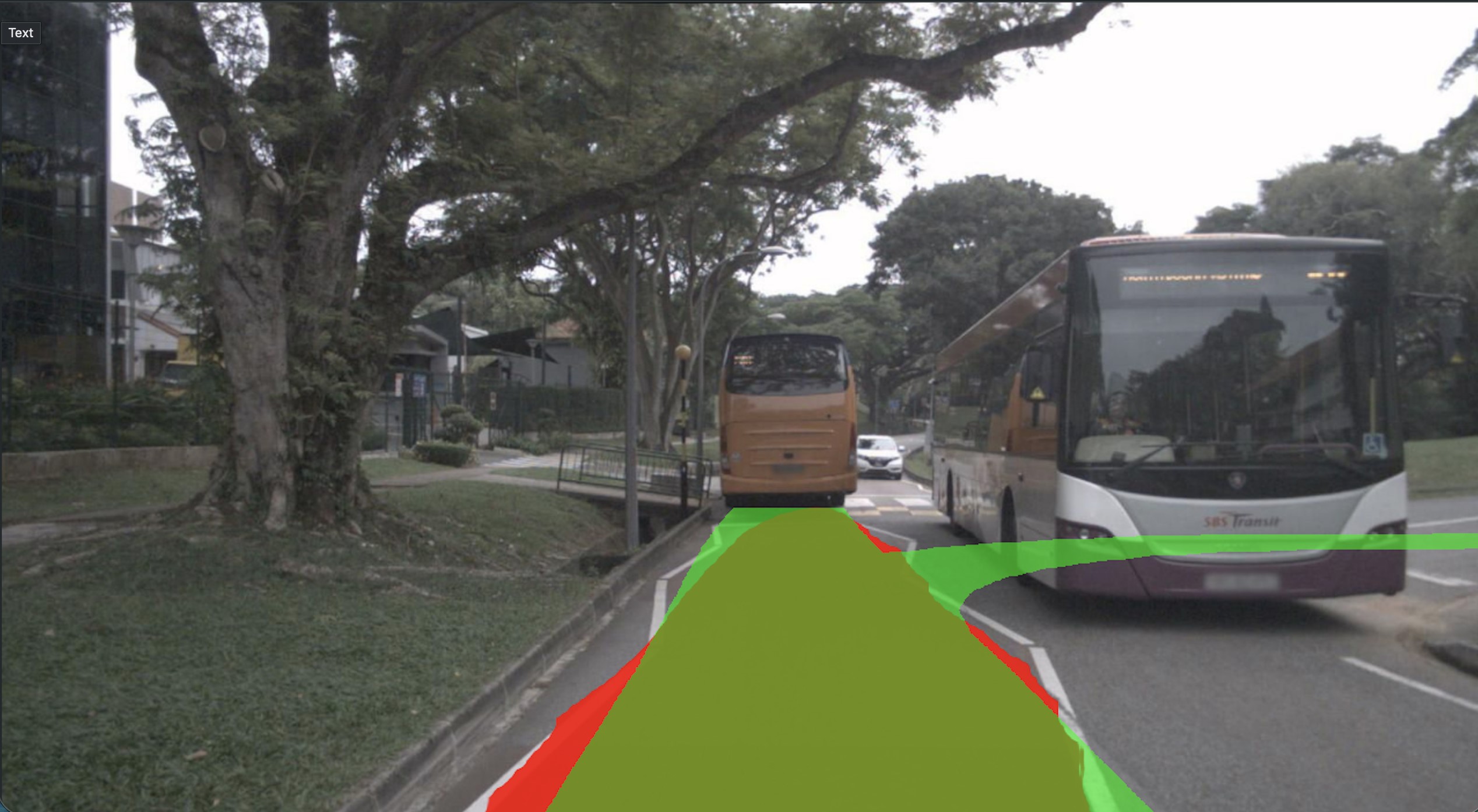}\label{fig:pretrain:fail_1_impossible_turn}}%
    \hfill
    \subfloat[Incoming Truck]{\includegraphics[width=0.24\linewidth]{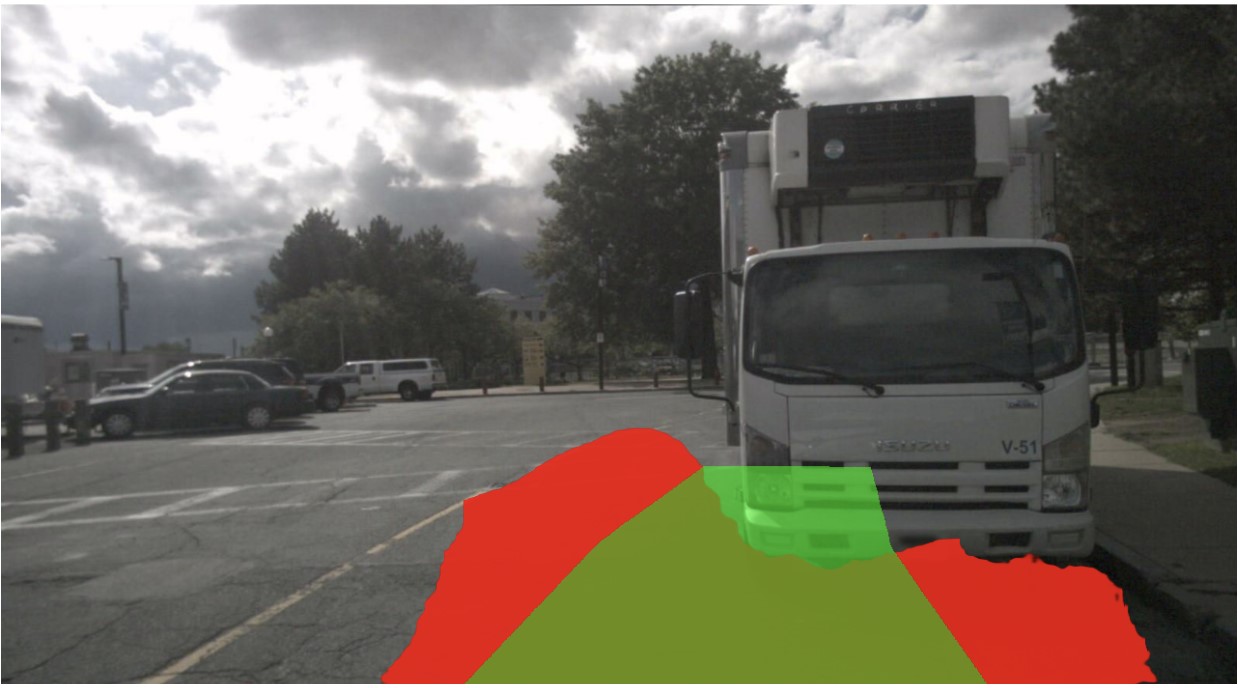}\label{fig:pretrain:Incoming_truck}}%
    \hfill
    \subfloat[GT Lane Misalignment]{\includegraphics[width=0.24\linewidth]{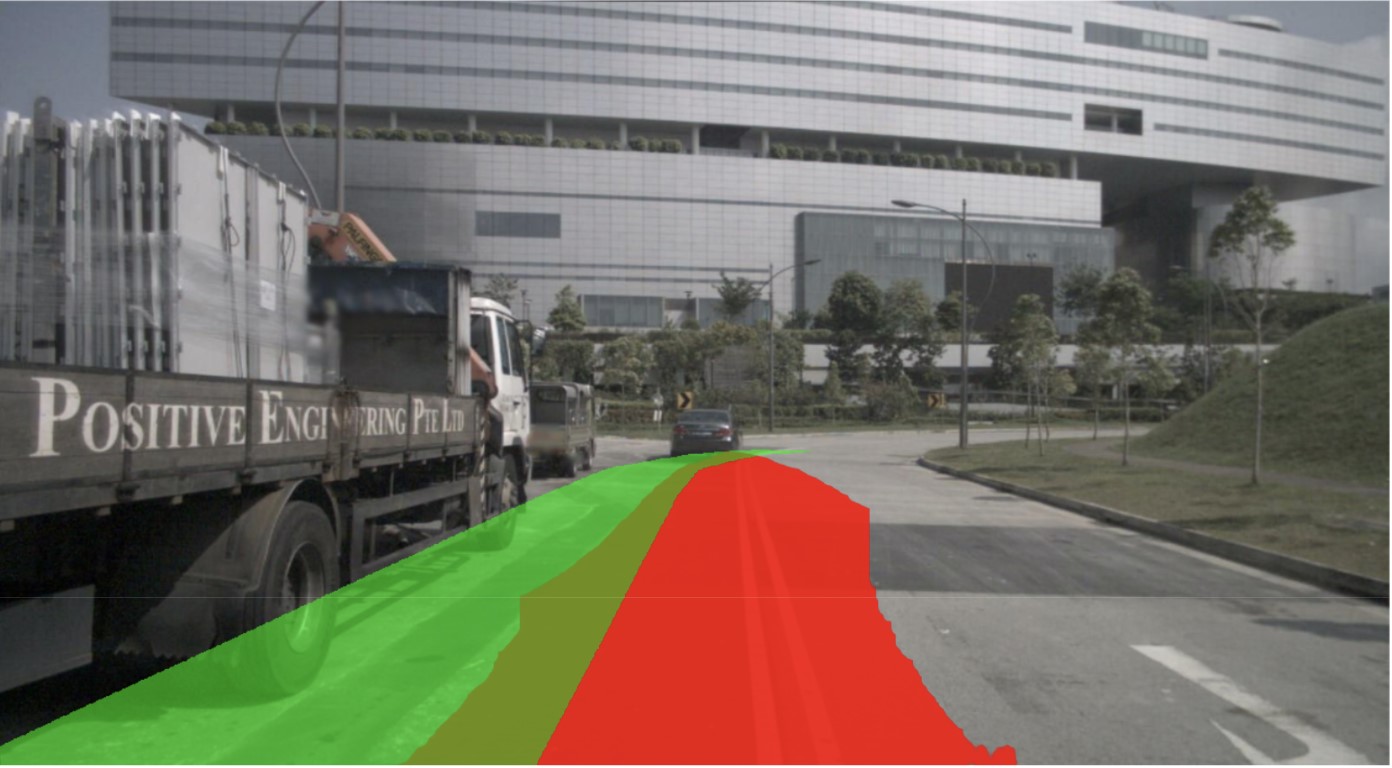}\label{fig:pretrain:GT_lane_misalignment}}%

    \caption{Pre-trained model predicted outputs shown in red and ground truth annotations in green.}
    \label{fig:pretrain_results}
\end{figure*}

\begin{figure}[t!]
    \centering
        \subfloat[FT: Bike Lane]{%
        \includegraphics[width=0.32\linewidth]{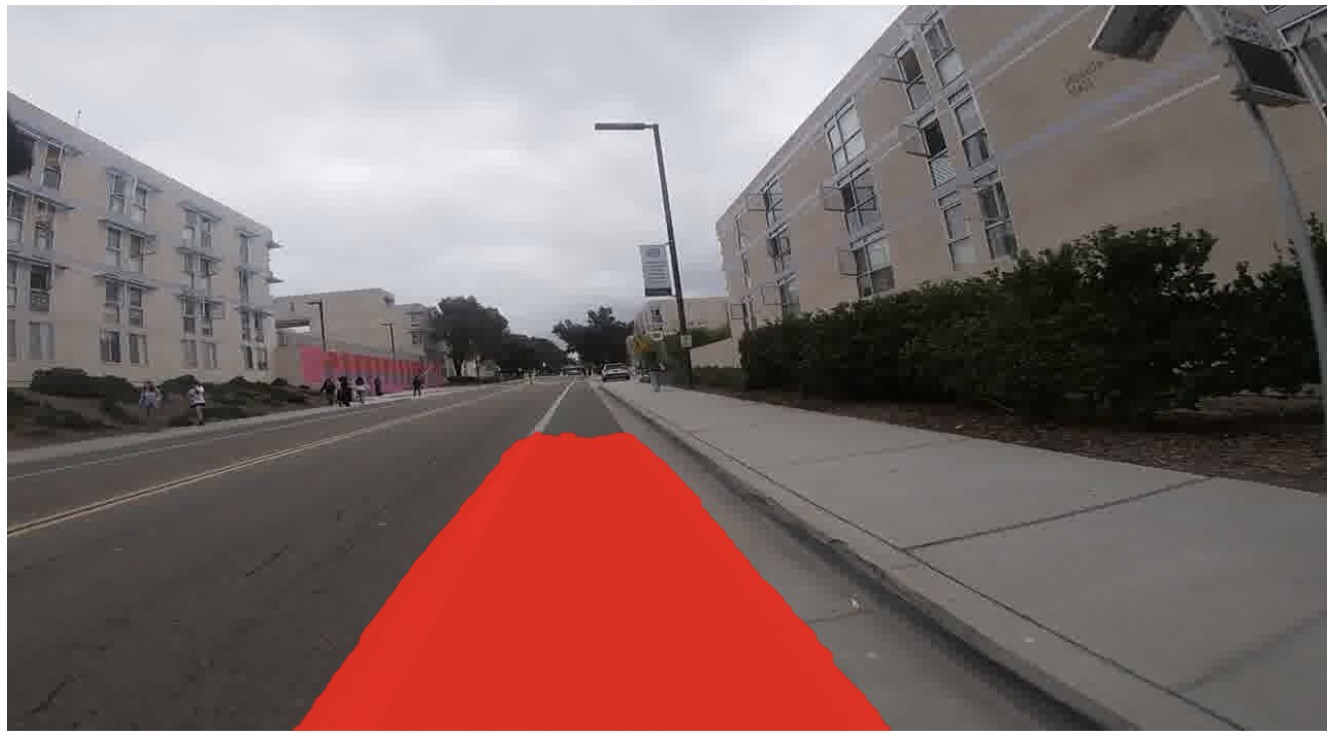}%
        \label{fig:scooter:bike_lane}
    }%
    \hfill
    \subfloat[FT: Shared Lane]{%
        \includegraphics[width=0.32\linewidth]{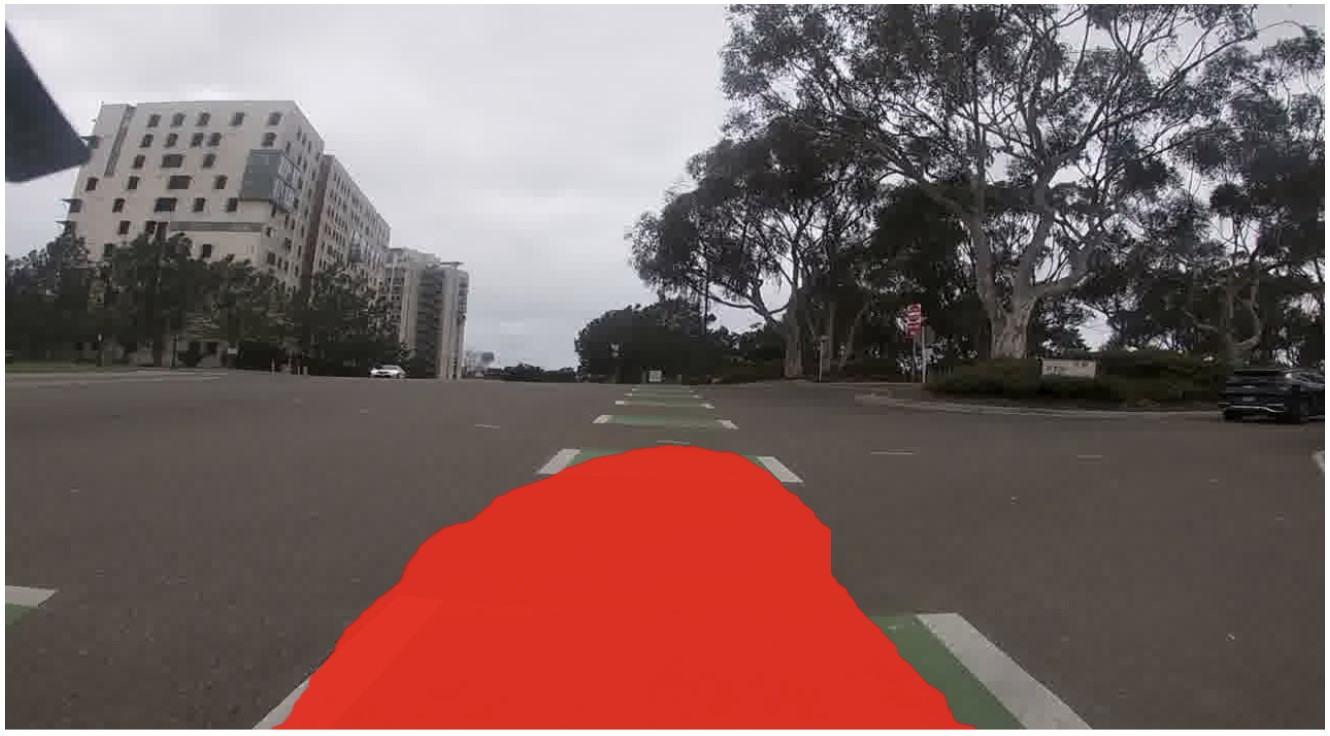}%
        \label{fig:scooter:shared_bike_lane}}%
    \hfill
    \subfloat[TS: Road]{%
        \includegraphics[width=0.32\linewidth]{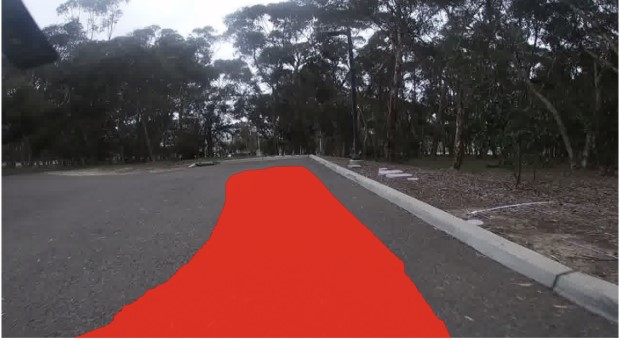}%
        \label{fig:scooter:scratch_road}}%

    \subfloat[FT: Road]{%
        \includegraphics[width=0.32\linewidth]{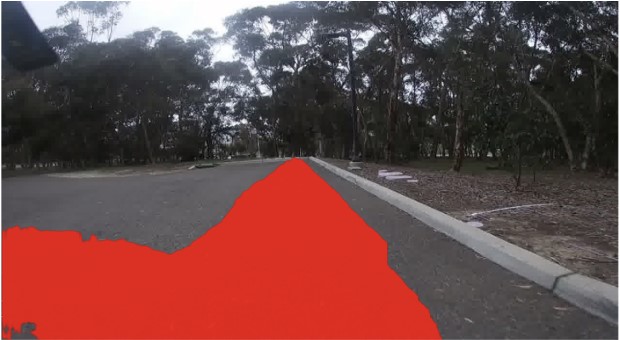}%
        \label{fig:scooter:ft_road}}%
    \hfill
    \subfloat[FT: Off-road]{%
        \includegraphics[width=0.32\linewidth]{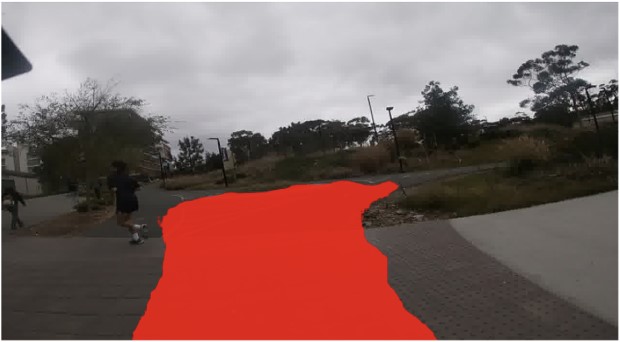}%
        \label{fig:scooter:ft_offroad}}%
    \hfill
    \subfloat[TS: Off-road]{%
        \includegraphics[width=0.32\linewidth]{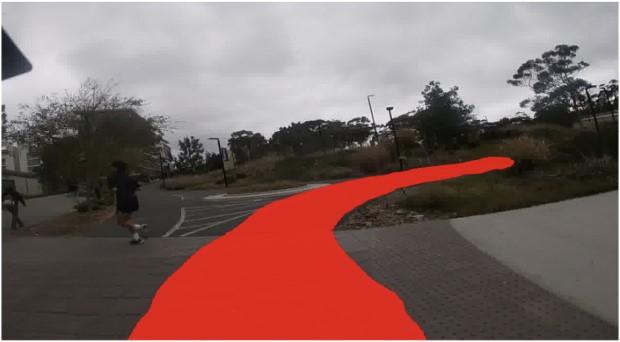}%
        \label{fig:scooter:scratch_offroad}}%
    \caption{Scooter predictions: Finetuned (FT) and Trained from Scratch (TS)}
    \label{fig:scooter_outputs}
\end{figure}

\subsection{Evaluation Setup}
For our evaluation, we generate ground truth trajectory segmentation masks from the OpenLaneV2~\cite{wang2023openlane} Subset B dataset built on NuScenes. We select this dataset because it provides a High Definition (HD) map for each scene which annotates lanes and lane connectivity information, which is essential for our reachability algorithm. For each frame, we first estimate the ego vehicle average speed over the next five seconds using future pose translations. From this speed, we calculate the distance the vehicle can cover over this time horizon. We then construct a directed lane graph from the provided lane centerlines and their connectivity matrix. To determine the current ego lane, we project the ego position into the lane coordinate frame and select the lane whose centerline is closest to the ego position. Starting from this lane, we traverse the lane graph while respecting the topology order and limit the traversal using the reachable distance budget. Only lanes that can be reached within this budget are retained. Then, each lane polyline is clipped to its reachable portion. The retained lanes therefore represent all feasible trajectory hypotheses within the five second horizon. For the perspective view projection and mask rasterization, we use the same method described in Section~\ref{subsec:pp}.

Once the ground truth masks are generated, we evaluate the predicted probabilistic masks using the Soft Intersection over Union metric (Ruzicka similarity~\cite{Ruzicka1958}). Let $G$ denote the ground truth mask and $P$ denote the predicted probability mask. The Soft IoU is defined as:
\begin{equation}
\label{eq:soft_iou}
    \text{Soft IoU} = 
    \frac{\sum_{i} \min(G_i, P_i)}
    {\sum_{i} \max(G_i, P_i)}
\end{equation}

We select this metric because our model outputs a probabilistic segmentation mask rather than a binary mask, and this formulation measures overlap without thresholding the predictions.

\subsection{Pre-training}
During pre-training, all images are resized to 704 by 1280, which is close to the 720p resolution we use to save image frames. The resolution is chosen because it preserves visual detail while ensuring that both spatial dimensions are divisible by 32, a requirement of the U-Net architecture due to its encoder downsampling stages. Images are normalized using standard ImageNet normalization to match the pretrained ResNet34 encoder weights. The model is trained using the Adam optimizer with a learning rate of $3\times 10^{-4}$ for 25 epochs. In the segmentation loss function (Equation~\ref{eq:loss}), we use an epsilon value of $0.1$, which effectively assigns nine times higher weight to false negatives compared to false positives, encouraging the model to prioritize the recall of positive regions.

\begin{table}[t]
  \centering
  \caption{Prediction Soft IoUs as defined by Equation~\ref{eq:soft_iou}}
  \label{tab:prediction_ious}
  \begin{tabular}{|l|c|c|}
    \hline
    \textbf{Model} & \textbf{Overall} & \textbf{Multi-Lane} \\
    \hline
    ResNet34 & 0.5579 & 0.5718 \\
    \hline
    SegFormer & 0.5645 & 0.5699 \\
    \hline
  \end{tabular}
\end{table}

Table~\ref{tab:prediction_ious} shows the quantitative results of the pretrained model on the OpenLaneV2 NuScenes split. For the multi-evaluation setting, after filtering the reachable lanes within the next five seconds in the ground truth, we select only those frames that contain at least four lanes. The presence of four lanes typically indicates a two-way road with a split in the road structure, ensuring that at least one alternative mode is available for evaluation. In addition to ResNet34, we also do the training and evaluation for SegFormer~\cite{xie2021segformer} backbone.

Fron the table, we can observe that both architectures perform very similarly, with only marginal gains when switching from a ResNet34-based vision backbone to the transformer-based SegFormer. The Soft IoU values lie in the 0.56–0.57 range, suggesting that the model captures a substantial portion of feasible future trajectories. While this leaves room for improvement, it is important to note that the evaluation itself is not perfect. The limitations of the HD-map-based ground truth and related edge cases are discussed in Section~\ref{subsec:lim-mask-gen}, along with qualitative examples. Another important observation is that the models do not collapse in multi-hypothesis scenarios (Overall v/s Multi-Lane columns of the table). In fact, the models either perform comparably or slightly better when multiple future lane options are present. This behavior supports our claim that the asymmetric loss formulation encourages multi-modal hypothesis coverage rather than convergence to a single narrow path.

Figure~\ref{fig:pretrain_results} shows qualitative predictions of the ResNet34 pretrained model on the OpenLaneV2 NuScenes split. In these images, the red mask indicates the predicted reachable region, while the green mask represents the ground truth generated using our evaluation setup. Figure~\ref{fig:pretrain:Straight} shows an example of straight driving, where our model succeeds in predicting the reachable area five seconds into the future. Figures~\ref{fig:pretrain:multimode_gt}, \ref{fig:pretrain:Singapore_Multimode}, and \ref{fig:pretrain:Singapore_Diversion} demonstrate that the model is able to predict multiple trajectory hypotheses, even in cases where the ground truth may not capture all possible modes, as discussed later in Section~\ref{subsec:lim-mask-gen}. The model also learns to avoid vehicles, as shown in Figure~\ref{fig:pretrain:Incoming_truck}, where one of the predicted modes avoids colliding with the incoming truck. Note that while not all hypotheses predicted by our model may be feasible due to dynamic agents in the scene, the model generally avoids clearly unfeasible hypotheses such as driving on the wrong side of the road or running into vehicles.

\subsection{Finetuning on Scooter driving videos}
We collected 40 minutes of driving videos from an electric scooter by attaching a GoPro HERO 7 White camera to it near the handle bar center. The collected data is from the UC San Diego campus, where we cover driving on roads, bike paths, and sidewalks. Among the 21k collected frames, we held out a test set of 1k frames and allocated the rest for training.

In our experiment, we finetuned the large-scale pre-trained model over the scooter data for 10 epochs. As a baseline, we also trained the model from scratch on the scooter data. The image and loss function setup remains the same as pre-training. While finetuning, we reduced the learning rate to $10^{-5}$ to prevent catastrophic forgetting and help the model adapt smoothly to the scooter domain.

Figure~\ref{fig:scooter_outputs} shows the qualitative predictions of the finetuned model and the model trained from scratch on the scooter data. Figures~\ref{fig:scooter:bike_lane} and ~\ref{fig:scooter:shared_bike_lane} show that the finetuned model has adapted to driving on regular and shared bike lanes. In addition, it also retains the pre-training imparted ability to predict multiple trajectory hypotheses when possible, as seen in Figures~\ref{fig:scooter:ft_road} and ~\ref{fig:scooter:ft_offroad}. In contrast, the model trained from scratch tends to predict a single hypothesis, possibly because it overfits the scooter data (Figures~\ref{fig:scooter:scratch_road} and \ref{fig:scooter:scratch_offroad}).
\section{Discussion \& Limitations}
\label{sec:discussion}
    \subsection{Geometric Constraints and Data Curation}
        While our pipeline processed approximately $10^6$ frames, valid trajectory masks were successfully recovered for roughly 650k frames. This discrepancy is primarily attributable to the highly unstructured and unconstrained nature of the source material. Operating on uncalibrated, monocular in-the-wild video presents significant challenges for traditional SfM frameworks. 
        
        The primary constraint is the degenerate motion profile inherent to automotive datasets: predominant forward-rectilinear movement causes extremely low parallax. Because SfM requires baseline diversity for robust triangulation, these near-zero parallax sequences represent a fundamental geometric limit of monocular reconstruction. However, given our dataset's massive scale, this expected data loss remains statistically inconsequential to the training objective. Furthermore, environmental factors, such as feature sparsity or dynamic objects violating the static-scene assumption, can induce erroneous trajectory divergence. While we readily filter out null frames, isolating geometrically corrupted masks remains challenging in an unsupervised setting. Future work could integrate temporal consistency checks or auxiliary monocular depth priors to mask dynamic objects and further purify the training signal.  
        
        Finally, while our methodology leverages a local planarity assumption during height calculation, we acknowledge that real-world environments frequently exhibit complex topological variations. Specifically, the planarity constraint may be transiently violated during high-gradient maneuvers, such as navigating highway interchanges or cresting hills. Though we adopt the median of all temporal height estimates as a robust global metric, sequences with many local planarity violations may still generate a skewed median height value.

    \subsection{Advantages Over Map-Dependent Architectures}
    
    \label{subsec:scale_ambiguity}
        Although monocular SfM recovers trajectories only up to scale, its deployment advantages extend beyond image-space segmentation. Compared to Bird's Eye View (BEV) architectures~\cite{vectornet, bevtp}, which depend on calibrated sensors or expensive HD maps, our method learns spatial priors directly from uncalibrated monocular data. When deployed on vehicles with calibrated sensors, the predicted perspective-view segmentation can be projected into 3D, enabling it to function as a robust, map-free BEV prior that supplies spatial context to downstream trajectory estimators and planners without requiring HD maps.
        
        Beyond BEV methods, many trajectory estimators use vectorized representations that model road topology and agents as graphs or polylines. While efficient and precise, these approaches rely heavily on accurate HD maps. Our segmentation-based method reduces this dependency by learning lane topology directly from vision, improving robustness in unmapped or dynamic environments.
        
        

    \subsection{Evaluation Validation}
    \label{subsec:lim-mask-gen}
        Evaluating our model on the NuScenes dataset provides a robust measure of its generalization capabilities because the data is entirely out-of-distribution compared to our training set. Specifically, NuScenes features driving sequences from Boston and Singapore, both of which were intentionally excluded during training. Furthermore, all the data in our training set is based on left-hand drive environments, whereas the Singapore data introduces right-hand drive scenarios. The NuScenes dataset also includes adverse weather conditions, such as rain, as shown in Figure~\ref{fig:pretrain:rainy}.
                
        Similar to \cite{barnes2017find}, we face the challenge of evaluating a segmentation-based formulation without the exact right type of ground truth data. While~\cite{barnes2017find} addresses this by overfitting a model to ego-trajectories in the evaluation set to establish a baseline, we instead leverage the HD maps provided by NuScenes. Moreover, because~\cite{barnes2017find} adopts a purely segmentation-based approach, it cannot effectively assess multi-modal predictions. By using the OpenLaneV2 framework, we overcome this limitation and quantitatively evaluate our model's ability to generate multiple feasible trajectory hypotheses simultaneously. 
        
        However, using HD maps for ground truth introduces structural limitations. First, the OpenLaneV2 HD map does not guarantee full topological connectivity between all legally reachable lanes. As shown in Figures~\ref{fig:pretrain:Singapore_Multimode} and \ref{fig:pretrain:Singapore_Diversion}, the assigned ego-lane can be disconnected from other valid trajectory lanes. As a result, our model may correctly predict a legally and physically feasible path yet be penalized because the map lacks the corresponding connection. Conversely, when the model predicts such missing links, it demonstrates genuine visual and physical reasoning rather than memorization of map structure.
        
        Second, HD maps are static and do not account for dynamic agents. Thus, ground truth may label trajectories as feasible even when blocked by vehicles or obstacles, as in Figures~\ref{fig:pretrain:fail_1_impossible_turn} and \ref{fig:pretrain:Incoming_truck}. This can penalize the model for correctly avoiding unsafe paths. At the same time, if the model suppresses a trajectory due to an observed obstacle despite static ground truth suggesting otherwise, its an evidence of dynamic scene awareness rather than reliance on static priors.
        
        Finally, the dataset includes maneuvers such as lane changes that are not reflected in the underlying HD map (Figure~\ref{fig:pretrain:GT_lane_misalignment}). In these cases, the default assigned ego-lane may not match the vehicle's true trajectory, introducing noise into evaluation and unfairly degrading precision and recall. This limitation highlights the rigidity and incompleteness of HD maps, reinforcing the motivation for our self-supervised, vision-based pipeline, which learns from real human driving behavior instead of brittle manual annotations.

\section*{ACKNOWLEDGMENT}

Portions of this manuscript were reviewed with the assistance of ChatGPT (OpenAI) and Gemini (Google) for the purposes of grammar correction, language refinement, and improving overall textual clarity. These tools were used solely to enhance readability and writing quality.

\bibliography{ref}
\bibliographystyle{IEEEtran}










\end{document}